\documentclass[10pt,twocolumn,letterpaper]{article}

\usepackage[pagenumbers]{cvpr} 

\renewcommand{\paragraph}[1]{\vspace{.5em}\noindent\textbf{#1}}

\usepackage{algorithm}
\usepackage[noend]{algpseudocode}
\newtheorem{lemma}{Lemma}
\newtheorem{proposition}{Proposition}

\newenvironment{proof}[1][Proof]{\par\noindent\textit{#1.}\ \ignorespaces}{\hfill$\square$\par\smallskip}
\DeclareMathOperator*{\argmin}{arg\,min}

\definecolor{cvprblue}{rgb}{0.21,0.49,0.74}
\usepackage[pagebackref,breaklinks,colorlinks,allcolors=cvprblue]{hyperref}

\def\confName{CVPR}
\def\confYear{2027}

\title{DoseSteer: Steering diffusion-based dose prediction models towards clinical targets via Dual-Response Guidance}

\author{%
Shupeng Cheng$^{1}$\quad
Jing Jia$^{2}$\quad
Jincheng Wu$^{1}$\quad
Zhangyu Wang$^{1}$\quad
Gao Wang$^{1}$\quad
Maohe Jiang$^{1}$\\
Chi Xu$^{3}$\quad
Feiyang Wang$^{1}$\quad
Zhilei Liu Shen$^{1}$\quad
Guanyang Wang$^{2}$\quad
Ruishan Liu$^{1}$\\[0.6em]
{\normalsize $^{1}$University of Southern California\qquad $^{2}$Rutgers University\qquad $^{3}$Rice University}%
}

\begin{document}
\maketitle
\begin{abstract}
AI dose-prediction models generate patient-specific 3D dose distributions from patient anatomy.
Yet these predictions can still miss structure-level clinical dose targets. Once deployed, most existing predictors offer no direct mechanism to correct such deviations, deferring target enforcement to downstream plan optimization.
We introduce DoseSteer, an inference-time framework that steers frozen diffusion-based dose prediction models towards specified clinical targets without retraining. To meet clinical targets while preserving overall dose prediction quality, we propose \textbf{D}ual-\textbf{R}esponse \textbf{G}uidance (DRG), a structure-centric algorithm that probes the model's responses to perturbations in each structure's noisy dose and context and jointly solves for per-structure steering strengths.
Across five dataset-model configurations spanning three diffusion architectures and three datasets, DRG outperforms 12 training-free baselines under identical reference-derived target proxies, removing at least 97\% of the target error, reducing the official dose error by 7.8\% on average, and improving dose fidelity outside the steered structures.
DoseSteer reconciles learned anatomical dose priors with explicit structure-level targets at test time, establishing a principled, training-free mechanism to refine AI dose predictions prior to plan optimization. Code is available at \url{https://github.com/csp2002/DoseSteer}.

\end{abstract}

\section{Introduction}
\label{sec:intro}

\begin{figure}[t]
\centering

\includegraphics[width=\linewidth]{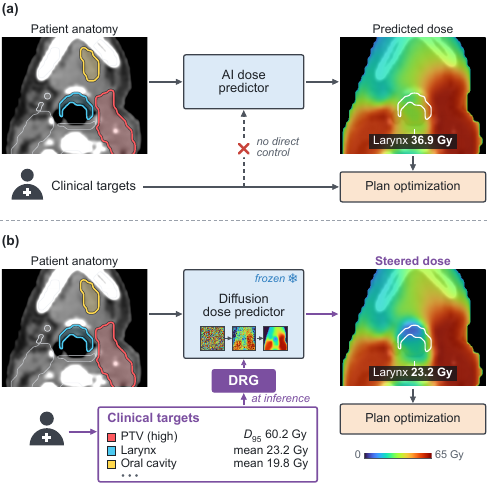}
\caption{\textbf{Motivation of DoseSteer. }
(a)~A dose predictor usually maps patient anatomy to a dose with no direct control during prediction, deferring clinical targets to downstream plan optimization.
(b)~In DoseSteer, clinical targets (e.g., PTV $D_{95}$ or OAR mean dose) enter at inference, where DRG steers a frozen diffusion predictor to resolve deviations (e.g., bringing larynx mean dose from 36.9 Gy to its 23.2 Gy target) prior to plan optimization.
}
\label{fig:teaser}
\end{figure}

In automated radiotherapy workflows, deep learning models predict patient-specific 3D dose distributions from patient anatomy~\cite{babier2021openkbp, wang2020review}.
Despite advances in average accuracy~\cite{kearney2018dosenet, nguyen20193d, kearney2020dosegan, gao2023flexible, wang2025radiotherapy}, AI-predicted dose can still miss structure-level clinical targets, such as exceeding an organ-at-risk (OAR) mean-dose limit or under-covering a planning target volume (PTV) (\cref{fig:teaser}). 
Since deployed predictors typically act as fixed black boxes, resolving these discrepancies is typically deferred to downstream plan optimization~\cite{babier2018knowledge, babier2022openkbp,mcintosh2017fully, zhang2021probabilistic,borderias2023machine,yu2025multicenter}.
We therefore explore a direction orthogonal to training a better model for higher raw prediction accuracy: use structure-level clinical targets to steer a trained dose predictor at inference,
thereby incorporating clinical targets directly into the AI prediction stage.

Recent conditional diffusion predictors~\cite{zhang2024dosediff, fu2024mddose, feng2023diffdp, cao2025beam, Wang_2026_CVPR} offer a natural handle for inference-time steering because their multi-step sampling trajectory allows progressive correction. At each denoising step, these networks refine a noisy 3D dose estimate conditioned on the planning image (CT or MRI) stacked with multi-channel anatomical representations such as structure masks and signed distance maps. 
While training-free guidance has flourished in natural image restoration and text-to-image synthesis, directly applying existing methods off the shelf overlooks the structure-centric nature of radiotherapy: clinicians specify treatment goals through discrete, structure-level dose objectives (\eg, OAR mean dose or PTV coverage), making \textbf{individual anatomical structures} the natural units of control.
Existing guidance methods struggle to align with this structure-centric nature in two ways~\cite{chung2023dps,ye2024tfg}. First, most methods manipulate only the noisy state (or clean estimate), treating the structure context as a \textit{static} background.
Second, they typically rely on a global guidance scale across all targets rather than solving for structure-specific strengths, making it difficult to coordinate multiple structures.

To address this, we introduce DoseSteer, an inference-time steering framework and its core algorithm, Dual-Response Guidance (DRG). Designed around anatomical structures as the natural unit of control, DRG steers through both the noisy state and the structure context of a frozen diffusion predictor and coordinates corrections across structures by jointly solving for per-structure steering strengths. Specifically, at each steering step, DRG probes the model to extract two complementary directions for each target structure: the \textbf{noise response} (by perturbing the structure's noisy dose) and the \textbf{context response} (by switching off its conditioning channels, analogous to structure-wise classifier-free guidance~\cite{ho2022cfg}). 
Rather than tuning step sizes by hand, DRG combines both responses across all target structures and jointly solves for all steering strengths in a single regularized least-squares step, adaptively balancing both levers across predictors and denoising steps.

We benchmark DRG against 12 training-free guidance methods across five dataset-model configurations spanning three diffusion architectures and three datasets (OpenKBP~\cite{babier2021openkbp}, GDP-HMM~\cite{gao2025gdphmm}, and an in-house cohort). Across all five configurations, DRG removes at least 97\% of the target error while reducing the official dose error by 7.8\% on average and achieving the highest dose fidelity outside the steered structures.
Our contributions are three-fold:
\begin{itemize}

\item \textbf{Inference-Time Dose Steering:} We introduce DoseSteer, a training-free framework to steer frozen diffusion dose predictors towards structure-level clinical targets. 

\item \textbf{Algorithm and Analysis:} We propose Dual-Response Guidance (DRG), which pairs structure-wise noise and context probing with a regularized joint solver across all target structures, alongside a theoretical explanation.

\item \textbf{Comprehensive Benchmarking:} We establish a unified evaluation across five dataset-model pairs and 12 baselines, demonstrating DRG's consistent gains in both target satisfaction and full-volume dose fidelity.
    
\end{itemize}

\begin{figure*}[t]
\centering
\includegraphics[width=\linewidth]{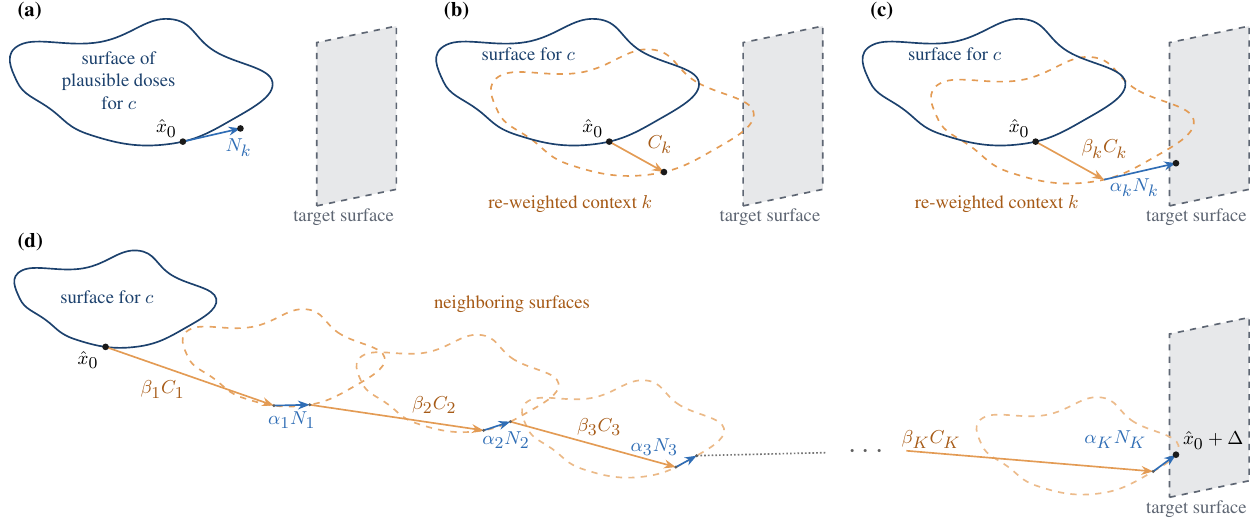}
\caption{\textbf{Geometric intuition for the two levers of DRG.} Solid curve: surface $\mathcal{M}_c$ of plausible doses under $c$; gray plane: target-satisfying surface.
(a)~The noise response $N_k\approx J\mathbf{1}_k$ moves $\hat{x}_0$ along $\mathcal{M}_c$.
(b)~The context response $C_k$ jumps towards a neighboring surface where structure $k$ is re-weighted.
(c, d)~Solving jointly for all $2K$ strengths combines context jumps across neighboring surfaces with within-surface noise steps to reach the target surface near the model's re-weighted belief (\cref{sec:analysis}).
Note that the tip-to-tail path in (d) is a geometric decomposition of the single simultaneous update $\Delta = \sum_k (\alpha_k N_k + \beta_k C_k)$, not a sequential structure-by-structure pass.}
\label{fig:levers}
\end{figure*}

\section{Related work}
\label{sec:related}

\subsection{Dose Prediction}
Deep learning approaches typically formulate 3D dose prediction as a deterministic image-to-image regression task using CNN or U-Net variants~\cite{kearney2018dosenet,nguyen20193d,liu2021cascade,nguyen2019feasibility,chen2019feasibility,barragan2019three} and Transformer-based architectures~\cite{jiao2023transdose,gheshlaghi2024cascade,wang2024latent}. Recently, conditional diffusion models~\cite{song2021ddim,ho2020ddpm} have been adopted to capture complex dose distributions conditioned on planning images and structure contexts~\cite{feng2023diffdp,fu2023sp,zhang2024dosediff,fu2024mddose,feng2024diffusion,liao2025fddm,cao2025beam,xie2025conditional,pan2025casmamdiff,Wang_2026_CVPR}. While several studies try to incorporate clinical preferences into AI models~\cite{gao2025preference,nguyen2020incorporating,nguyen2019generating,bohara2020using,jensen2021novel,ma2021feasibility,Wang_2026_CVPR}, they usually require specialized training and additional preference-labeled data. Steering frozen dose predictors to satisfy clinical dose targets during inference is still underexplored.

\subsection{Training-Free Guidance for Diffusion Models}
Steering a pre-trained diffusion model to satisfy external constraints without retraining broadly falls into three categories. First, clean-space correction and alternating inference methods directly edit the denoiser's clean estimate $\hat{x}_{0}$ via measurement-consistent projection, proximal updates, or alternating clean-space sampling and re-noising~\cite{wang2023ddnm, zhu2023diffpir,kawar2022denoising,zhang2025daps,kim2026ddiff}. Second, gradient-based guidance methods steer sampling using gradients of a target objective with respect to the noisy state $x_{t}$ or clean estimate $\hat{x}_{0}$~\cite{chung2022mcg,chung2023dps,bansal2023universal,yu2023freedom,song2023loss,he2024mpgd,ye2024tfg}, often preconditioned via Tweedie posterior covariance approximations~\cite{song2023pigdm,boys2024tmpd,efron2011tweedie,kadkhodaie2021stochastic,manor2024posterior,meng2021highorder,robbins1956empirical}. Third, noise optimization and sample selection methods optimize sampling noise trajectories or resample across parallel candidates~\cite{tang2025dno,wallace2023end,singhal2025fk}. These methods usually treat the conditioning context as a fixed background and rely on a global guidance scale to guide generation. By contrast, DRG leverages both the multi-channel structure context and the noisy state as active structure-level steering levers and jointly solves for structure-specific steering strengths to meet clinical targets while preserving overall dose prediction quality.

\section{Method}
\label{sec:method}

Dual-Response Guidance (DRG) steers a trained conditional diffusion dose predictor, without retraining, so that $K$ structure-level dose metrics meet clinical targets.
At each steering step it corrects the clean estimate with the model's own responses to its two inputs, the noisy state and the structure context (\cref{sec:levers,sec:drg}); \cref{sec:prelim} recalls the preliminaries and \cref{sec:analysis} analyzes the algorithm.

\subsection{Preliminaries}
\label{sec:prelim}

A dose predictor maps a conditioning context $c$ to a dose $x_0\in\mathbb{R}^n$ on $n$ voxels in normalized model units (affine in Gy).
The context $c$ stacks the planning image with per-structure conditioning channels (masks and, for some models, signed distance maps); setting structure $k$'s channels to zero yields $c_{\backslash k}$ as an approximation of removing structure $k$, with $c_k = c - c_{\backslash k}$.
Diffusion predictors such as DoseDiff~\cite{zhang2024dosediff} and MD-Dose~\cite{fu2024mddose} model $p(x_0\mid c)$ with the forward process $x_t=\sqrt{\bar\alpha_t}\,x_0+\sqrt{1-\bar\alpha_t}\,\varepsilon$, $\varepsilon\sim\mathcal{N}(0,I)$ and a noise schedule $\bar\alpha_t$ decreasing in $t$~\cite{ho2020ddpm}.
A model $\epsilon_\theta(x_t,c)$ predicts the noise and defines the clean estimate, Tweedie's estimate of $x_0$ (\cref{lem:tweedie}),
\begin{equation}\label{eq:x0}
\hat{x}_0=f_\theta(x_t,c)=\frac{x_t-\sqrt{1-\bar\alpha_t}\,\epsilon_\theta(x_t,c)}{\sqrt{\bar\alpha_t}}.
\end{equation}
Sampling runs deterministic DDIM~\cite{song2021ddim} over $T$ steps, $t=T-1,\dots,0$, with $\bar\alpha_{-1}=1$.
We write a step with the clean estimate and the noise consistent with it,
\begin{equation}\label{eq:ddim}
\begin{gathered}
\hat\varepsilon=\frac{x_t-\sqrt{\bar\alpha_t}\,\hat{x}_0}{\sqrt{1-\bar\alpha_t}},\\
x_{t-1}=\sqrt{\bar\alpha_{t-1}}\,\hat{x}_0+\sqrt{1-\bar\alpha_{t-1}}\,\hat\varepsilon,
\end{gathered}
\end{equation}
which is plain DDIM for $\hat{x}_0$ from \cref{eq:x0}, lets a corrected $\hat{x}_0$ enter directly and returns the dose $\hat{x}_0$ at $t=0$.

\makeatletter\par\penalty-100\begingroup\dimen@=\pagegoal\advance\dimen@-\pagetotal\ifdim\dimen@<4\baselineskip\newpage\fi\endgroup\makeatother
\begin{lemma}[Tweedie's identities~\cite{robbins1956empirical,efron2011tweedie,kadkhodaie2021stochastic,meng2021highorder,manor2024posterior}]\label{lem:tweedie}
For any $p(x_0\mid c)$, the posterior mean and covariance are
\[
\begin{gathered}
\mathbb{E}[x_0\mid x_t,c]=\frac{x_t-\sqrt{1-\bar\alpha_t}\,\varepsilon^\star}{\sqrt{\bar\alpha_t}},\\
\operatorname{Cov}[x_0\mid x_t,c]=\kappa_t\,\frac{\partial\,\mathbb{E}[x_0\mid x_t,c]}{\partial x_t}, \qquad\kappa_t=\frac{1-\bar\alpha_t}{\sqrt{\bar\alpha_t}},
\end{gathered}
\]
where $\varepsilon^\star=\mathbb{E}[\varepsilon\mid x_t,c]$ minimizes the noise-prediction loss.
\end{lemma}

\noindent We assume \textbf{(A0)} that the denoiser attains this minimum, $f_\theta(\cdot,c)=\mathbb{E}[x_0\mid x_t=\cdot\,,c]$; its Jacobian is then a scaled posterior covariance, $\Sigma_t:=\operatorname{Cov}[x_0\mid x_t,c]=\kappa_tJ$.

\subsection{Probing the two responses}
\label{sec:levers}

\paragraph{Two inputs.}
Training-free steering can act on $f_\theta$ through its two inputs, $x_t$ and $c$.
To first order, perturbing both shifts the clean estimate by
\begin{equation}\label{eq:twoinputs}
\delta\hat{x}_0=J\,\delta x_t+J_c\,\delta c,\qquad J=\frac{\partial f_\theta}{\partial x_t},\qquad J_c=\frac{\partial f_\theta}{\partial c} .
\end{equation}
Existing guidance typically uses only the first term (with $J$ or an identity approximation) while keeping $c$ fixed.
DRG activates both terms structure by structure: it probes $J\mathbf{1}_k$ via a finite difference on structure $k$'s noisy dose (the \textbf{noise response}), and probes $J_c c_k$ via a chord between $c_{\backslash k}$ and $c$ (the \textbf{context response}).  

\paragraph{Noise response.}
Let $\Omega_k$ denote the voxels of structure $k$, and let $\mathbf{1}_k \in \{0,1\}^n$ indicate its active target voxels: all of $\Omega_k$ for an OAR, or the coldest 10\% of $\Omega_k$ at the current $\hat{x}_0$ for a PTV (updated at each steering step).
Raising $x_t$ along $\mathbf{1}_k$ yields the noise response via central finite differences with step $h$,
\begin{equation}\label{eq:noise}
N_k=\frac{f_\theta(x_t+h\mathbf{1}_k,c)-f_\theta(x_t-h\mathbf{1}_k,c)}{2h}\approx J\mathbf{1}_k .
\end{equation}
Under (A0), $J\mathbf{1}_k$ is, up to $\kappa_t$, the covariance of structure $k$'s dose with the dose everywhere: how the model expects the whole dose to move with it, inside the structure and beyond.
Under the manifold hypothesis~\cite{chung2022mcg,he2024mpgd}, the estimates lie on or near a low-dimensional surface $\mathcal{M}_c$ of plausible doses, and a push of $x_t$ moves $\hat{x}_0$ along it (\cref{fig:levers}a; Supplementary Material, hereafter Supp.\ for simplicity).

\paragraph{Context response.}
The noise lever, a push on $x_t$, moves $\hat{x}_0$ only where the posterior varies: where the context pins structure $k$'s dose, $\Sigma_t\mathbf{1}_k\approx0$ and no cheap push on $x_t$ reaches a distant target; the context lever, a change of $c$, changes what it pins.
Switching off $c_k$ gives a chord of $J_c$,
\begin{equation}\label{eq:chord}
\begin{aligned}
D_k&=f_\theta(x_t,c)-f_\theta(x_t,c_{\setminus k})\\
&=\textstyle\int_0^1J_c(x_t,c_{\setminus k}+s\,c_k)\,c_k\,ds .
\end{aligned}
\end{equation}
DRG extends the chord linearly to a weight $w_k=1+\beta_k$ on the structure, $\hat{x}_0+\beta_kD_k$, and keeps it local, $C_k=M_k\odot D_k$, with $M_k$ a linear feather, $1$ on structure $k$ and $0$ beyond a distance $d=20$\,mm.
Where $M_k=1$, $\hat{x}_0+\beta_kC_k=f_\theta(x_t,c_{\setminus k})+w_kD_k$: classifier-free guidance~\cite{ho2022cfg} on structure $k$ alone.
It moves $\hat{x}_0$ towards the model's estimate when structure $k$'s context counts $w_k$ times (\cref{prop:responses}), near the neighboring surface of that context (\cref{fig:levers}b).
Notably, the network is evaluated only at $c$ and $c_{\setminus k}$, never at continuously re-weighted contexts.

\subsection{Dual-response guidance}
\label{sec:drg}

\paragraph{Targets and joint strengths.}
DRG receives scalar targets $y_k$ for $K$ structures, specifying either the mean dose of an OAR or the $D_{95}$ of a PTV (the minimum dose received by 95\% of its voxels). Let $q_k(\hat{x}_0)$ denote this clinical metric and $r_k = y_k - q_k(\hat{x}_0)$ its residual. To solve for the correction linearly, we define $m_k(x) = \mathbf{1}_k^\top x / \vert{}\mathbf{1}_k\vert{}$, which equals $q_k$ for an OAR and averages the coldest 10\% of voxels as a smooth linear surrogate for PTV $D_{95}$ (exact under a uniform shift). At each steering step $t \in \mathcal{S}$, DRG updates $\hat{x}_0 \leftarrow \hat{x}_0 + \Delta$ using a linear combination of all $2K$ responses (\cref{alg:drg}):
\begin{equation}\label{eq:delta}
\Delta = \sum_{k=1}^K \alpha_k N_k + \sum_{k=1}^K \beta_k C_k. 
\end{equation}
By linearity of $m$, $\Delta$ changes the surrogate metrics by $m(\Delta)=G^N\alpha+G^C\beta=G\vartheta$, with strengths $\vartheta=(\alpha,\beta)$, $G^N_{jk}=m_j(N_k)$, $G^C_{jk}=m_j(C_k)$ and $G=[G^N\;G^C]$, so the surrogate target-matching equations are $G\vartheta=r$.

\begin{algorithm}[t]
\caption{Dual-Response Guidance (DRG)}
\label{alg:drg}
\small
\algrenewcommand\algorithmicindent{1em}
\begin{algorithmic}
\Require $f_\theta$, context $c$, targets $y_1,\dots,y_K$, steering steps $\mathcal{S}$
\State $x_{T-1}\sim\mathcal{N}(0,I)$
\For{$t=T-1,\dots,0$}
  \State $\hat{x}_0\gets f_\theta(x_t,c)$ \Comment{clean estimate}
  \If{$t\in\mathcal{S}$}
    \For{$k=1,\dots,K$}
      \State $r_k\gets y_k-q_k(\hat{x}_0)$ \Comment{distance to the target}
      \State $N_k\gets\frac{f_\theta(x_t+h\mathbf{1}_k,c)-f_\theta(x_t-h\mathbf{1}_k,c)}{2h}$ \Comment{noise}
      \State $C_k\gets M_k\odot\big[f_\theta(x_t,c)-f_\theta(x_t,c_{\setminus k})\big]$ \Comment{context}
    \EndFor
    \State $(\alpha,\beta)\gets\argmin_{|\beta_k|\le B}\Phi(\alpha,\beta)$ \Comment{\cref{eq:solve}}
    \State $\hat{x}_0\gets\hat{x}_0+\sum_k\alpha_kN_k+\sum_k\beta_kC_k$
  \EndIf
  \State $\hat\varepsilon\gets\frac{x_t-\sqrt{\bar\alpha_t}\,\hat{x}_0}{\sqrt{1-\bar\alpha_t}}$ \Comment{noise consistent with $\hat{x}_0$}
  \State $x_{t-1}\gets\sqrt{\bar\alpha_{t-1}}\,\hat{x}_0+\sqrt{1-\bar\alpha_{t-1}}\,\hat\varepsilon$ \Comment{DDIM step}
\EndFor
\State \Return $\hat{x}_0$
\end{algorithmic}
\end{algorithm}

\paragraph{Pricing the pushes.}
Both responses are linear readings of the model, so the corrected estimate stays close to the mean of the model's own re-weighted belief, and hence plausible, as long as the pushes are \textit{small} (\cref{prop:tilt,prop:responses,prop:surface}).
The two levers are pushed in different units, a dose added to $x_t$ and a weight on a structure's context, so ``small'' needs a common currency, the price
\begin{equation}\label{eq:price}
\mathcal{P}(\vartheta)=\sum_k\Big(\frac{\lVert G^N_{:,k}\alpha_k\rVert^2}{v_k}+\frac{\beta_k^2}{\sigma_\beta^2}\Big).
\end{equation}
For the noise lever, \cref{lem:tweedie} under (A0) with $N_k=J\mathbf{1}_k$ gives the model's uncertainty about $m_k$ from the noise response itself,
\begin{equation}\label{eq:vk}
v_k=\operatorname{Var}[m_k(x_0)\mid x_t,c]=\frac{\kappa_t\,G^N_{kk}}{|\mathbf{1}_k|} ,
\end{equation}
so the first term is the squared change that the push on structure $k$ causes in all $K$ metrics, in units of the posterior variance of $m_k$: a structure whose dose the model is sure about is expensive to push, one it is unsure about is cheap.
For the context lever, $\beta_k=w_k-1$ is a change of a guidance weight, priced in units of $\sigma_\beta$ and bounded by $B$; $\sigma_\beta$, chosen per model on validation data, is the one knob that trades the two levers against each other.

\paragraph{Shape term.}
Because a target constrains only one summary statistic per structure, $\mathcal{V}(\vartheta) = \sum_k \text{Var}_{\Omega_k}(\Delta)$ penalizes intra-structure dose variations so corrections shift rather than distort the predicted profile.

\paragraph{Solve.}
DRG takes the minimizer over $|\beta_k|\le B$ of
\begin{equation}\label{eq:solve}
\Phi(\vartheta)=\lVert G\vartheta-r\rVert^2+\lambda\,\mathcal{V}(\vartheta)+\mu\,\mathcal{P}(\vartheta),
\end{equation}
one damped Gauss--Newton step on the $K$ targets~\cite{nocedal2006numerical}, with the damping set from the responses themselves (Supp.),
\begin{equation}\label{eq:mu}
\mu=\frac{10^{-3}}{2K}\Big(\sum_kv_k+\sigma_\beta^2\sum_k\lVert G^C_{:,k}\rVert^2\Big).
\end{equation}
Like $\sigma_\beta$, $\lambda$ is chosen per model on validation data.

\subsection{Analysis}
\label{sec:analysis}

Under the ideal-denoiser approximation (A0), Propositions 1 to 3 interpret DRG's two levers as linear approximations of posterior re-weightings (proofs in Supp.).

\begin{proposition}[Both levers re-weight the posterior]\label{prop:tilt}
For any $p(x_0\mid c)$, every push $\delta$ and every structure $k$,
\begin{equation*}
\begin{gathered}
p(x_0\mid x_t+\delta,c)\propto p(x_0\mid x_t,c)\,e^{a^\top x_0},\quad a=\frac{\delta}{\kappa_t},\\
p(x_0\mid x_t,c)\propto p(x_0\mid x_t,c_{\setminus k})\,e^{\ell_k(x_0)},\\
\ell_k(x_0)=\log p(c_k\mid x_0,c_{\setminus k}),
\end{gathered}
\end{equation*}
where $\ell_k(x_0)$ is the log-likelihood of structure $k$'s contour when the dose is $x_0$.
\end{proposition}
\noindent In this reading, neither lever invents a dose: a push on $x_t$ re-weights the doses the model already believes in, and a structure's context re-weights the same belief by how likely its contour is for each dose.

\begin{proposition}[The two responses under (A0)]\label{prop:responses}
Let $\delta = \sum_k \alpha_k \mathbf{1}_k$ be the net noise push on $x_t$, and let $\bar{x}(w)$ be the mean of $q_w\propto p(x_0\mid x_t,c_{\setminus k})\,e^{w\ell_k(x_0)}$, the posterior in which structure $k$'s contour counts $w$ times.
Under (A0) at $c$ and $c_{\setminus k}$, $\hat{x}_0=\bar{x}(1)$, $D_k=\bar{x}(1)-\bar{x}(0)$, and
\begin{equation*}
\begin{aligned}
\textup{(i)}&\quad f_\theta(x_t+\delta,c)=\hat{x}_0+J\delta+R(\delta),\\
\textup{(ii)}&\quad \bar{x}(1+\beta_k)=\hat{x}_0+\beta_kD_k+R_k(\beta_k),
\end{aligned}
\end{equation*}
where $R(\delta)=O(\lVert\delta\rVert^2)$ vanishes for a Gaussian posterior and $R_k$ vanishes when $\bar{x}(w)$ is affine in $w$.
\end{proposition}
\noindent The two responses thus approximate the two re-weightings linearly: $\sum_k\alpha_kN_k\approx J\delta$ captures the first-order push, and $\beta_kD_k$ extends the chord between the two contexts to the weight $w_k=1+\beta_k$; both remainders are bounded in Supp.

\begin{proposition}[DRG's estimate and the model's belief]\label{prop:surface}
Let $\bar{x}$ be the mean of $q\propto p(x_0\mid x_t,c)\,e^{a^\top x_0+\sum_k\beta_k\ell_k(x_0)}$, the posterior re-weighted by the push and by the $K$ contour likelihoods.
Under (A0) at $c$ and every $c_{\setminus k}$,
\begin{multline*}
\big\lVert\hat{x}_0+\Delta-\bar{x}\big\rVert\le\lVert R(\delta)\rVert+\textstyle\sum_k|\alpha_k|\,e_k\\
+\textstyle\sum_k|\beta_k|\big(\lVert(1-M_k)\odot D_k\rVert+\rho_k\big)+E(\delta,\beta),
\end{multline*}
with $e_k=\lVert N_k-J\mathbf{1}_k\rVert$ the finite-difference error, $|\beta_k|\rho_k$ a bound on $\lVert R_k(\beta_k)\rVert$, and $E$ second order in the two pushes.
\end{proposition}
\noindent Every term vanishes with the pushes, which is why the solve prices them: for small pushes, the corrected estimate stays close to the mean of the model's belief after both re-weightings, the geometric sense in which DRG reaches the target surface along plausible surfaces (\cref{fig:levers}d).

Because Propositions 1 to 3 rely on the ideal-denoiser approximation (A0), linear approximation, and zeroed channels as a proxy for structure removal, they serve as principled design intuition rather than strict finite-sample guarantees for trained networks.

\section{Experiments}
\label{sec:experiments}

\subsection{Setup}
\label{sec:setup}

\begin{table*}[t]
\centering
\caption{\textbf{Dataset statistics.}
Dataset splits, and the steered structures of a test plan: OARs, PTVs and their total $K$, as range (median).}
\label{tab:settings}
\footnotesize
\setlength{\tabcolsep}{7pt}
\begin{tabular}{@{}llllccccc@{}}
\toprule
\textbf{Dataset} & \textbf{Source} & \textbf{Site} & \textbf{Technique} & \textbf{Image} & \textbf{Train / val / test plans} & \textbf{OARs} & \textbf{PTVs} & \textbf{\boldmath$K$} \\
\midrule
OpenKBP & Challenge & Head and neck & IMRT & CT & 200 / 40 / 100 & 1--7 (6) & 1--3 (2) & 2--9 (8) \\
GDP-HMM & Challenge & Head and neck, lung & IMRT, VMAT & CT & 2727 / 148 / 356 & 8--27 (17) & 1--3 (2) & 9--28 (19) \\
Local & In-house & Brain & GK & MR & 101 / 16 / 38 & 2--5 (2) & 1--3 (1) & 3--7 (3) \\
\bottomrule
\end{tabular}
\end{table*}

\begin{table*}[t]
\centering
\caption{\textbf{Main quantitative comparison across five settings.}
Percentage error reduction relative to unsteered predictions (higher is better) on OpenKBP with DoseDiff (DD), MD-Dose (MD), and DoseDiT (DiT), GDP-HMM with DoseDiff (GDP), and the local cohort with DoseDiff (Local). All methods share identical targets, initial noise, and validation rules. Best per column in bold (target-error column unbolded as direct edit, DDNM, and DiffPIR enforce targets directly to $\ge 99.9\%$, while DRG achieves $\ge 97.4\%$).
}
\label{tab:main}
\footnotesize
\setlength{\tabcolsep}{2.2pt}
\begin{tabular}{lcccccccccccccccccccc}
\toprule
& \multicolumn{5}{c}{\textbf{Target error}} & \multicolumn{5}{c}{\textbf{Official metric}} & \multicolumn{5}{c}{\textbf{DVH score}} & \multicolumn{5}{c}{\textbf{Outside error}} \\
\cmidrule(lr){2-6} \cmidrule(lr){7-11} \cmidrule(lr){12-16} \cmidrule(lr){17-21}
\textbf{Method} & \textbf{DD} & \textbf{MD} & \textbf{DiT} & \textbf{GDP} & \textbf{Local} & \textbf{DD} & \textbf{MD} & \textbf{DiT} & \textbf{GDP} & \textbf{Local} & \textbf{DD} & \textbf{MD} & \textbf{DiT} & \textbf{GDP} & \textbf{Local} & \textbf{DD} & \textbf{MD} & \textbf{DiT} & \textbf{GDP} & \textbf{Local} \\
\midrule
\multicolumn{21}{c}{\textit{Clean-space correction and alternating inference}} \\
\midrule
Direct edit & 100.0 & 100.0 & 100.0 & 100.0 & 100.0 & 0.8 & $-0.5$ & 1.0 & 3.6 & 3.0 & 40.5 & 42.9 & 43.5 & 43.8 & $\mathbf{44.7}$ & 0.0 & 0.0 & 0.0 & 0.0 & 0.0 \\
DDNM~\cite{wang2023ddnm} & 99.9 & 99.9 & 100.0 & 100.0 & 100.0 & 1.0 & $-0.3$ & 2.6 & 8.9 & 2.2 & 42.4 & 42.1 & 45.9 & 50.0 & 42.4 & 0.4 & 0.7 & 1.8 & 4.2 & $-2.0$ \\
DiffPIR~\cite{zhu2023diffpir} & 99.9 & 99.9 & 100.0 & 100.0 & 100.0 & 1.1 & $-0.3$ & 2.6 & 9.8 & 2.3 & 42.4 & 41.8 & 45.3 & 49.1 & 43.1 & 0.5 & 0.7 & 1.7 & 5.0 & $-1.8$ \\
DAPS~\cite{zhang2025daps} & 9.6 & 10.3 & 13.8 & 1.3 & $-1.0$ & $-2.7$ & $-2.9$ & $-1.9$ & $-2.1$ & $-8.4$ & 5.7 & 3.7 & 7.7 & 3.0 & 5.2 & $-2.7$ & $-2.6$ & $-1.7$ & $-2.1$ & $-5.1$ \\
DDiff~\cite{kim2026ddiff} & 29.9 & 30.7 & 38.8 & 15.8 & 12.8 & 0.0 & 1.3 & 2.0 & 3.0 & 0.7 & 7.8 & 10.3 & 10.6 & 10.1 & 7.2 & $-0.4$ & 0.9 & 1.6 & 1.4 & 1.4 \\
\midrule
\multicolumn{21}{c}{\textit{Gradient-based guidance}} \\
\midrule
DPS~\cite{chung2023dps} & 27.5 & 35.6 & 30.4 & 3.2 & 12.2 & 0.2 & 1.0 & 1.3 & 4.8 & $-0.1$ & 10.5 & 11.3 & 11.0 & $-8.8$ & 6.6 & $-0.1$ & 0.5 & 0.9 & 2.6 & $-0.9$ \\
$\Pi$GDM~\cite{song2023pigdm} & 56.0 & 63.3 & 69.5 & 78.9 & 44.0 & 0.7 & 1.3 & 1.9 & 9.2 & 2.3 & 27.7 & 35.2 & 33.9 & 40.3 & 21.0 & 0.0 & 0.3 & 1.2 & 4.5 & $-0.1$ \\
TMPD~\cite{boys2024tmpd} & 21.0 & 67.0 & 32.5 & 51.8 & 3.5 & 0.8 & 3.7 & 2.0 & 6.9 & 0.2 & 7.3 & 36.3 & 17.3 & 29.0 & 1.1 & 0.6 & 2.9 & 1.6 & 4.1 & $-0.6$ \\
TFG~\cite{ye2024tfg} & 43.7 & 34.6 & 41.4 & 18.3 & 42.2 & 0.5 & 1.2 & 1.7 & 4.5 & 0.0 & 5.2 & 7.1 & 2.4 & 4.5 & 17.2 & 0.0 & 0.4 & 1.0 & 2.4 & $-1.0$ \\
\midrule
\multicolumn{21}{c}{\textit{Noise optimization and sample selection}} \\
\midrule
DNO~\cite{tang2025dno} & 18.6 & 51.2 & 29.7 & 84.1 & 17.9 & 0.3 & 3.5 & 2.3 & 10.2 & 0.8 & 10.3 & 31.2 & 18.0 & 44.5 & 8.3 & 0.1 & 3.1 & 1.7 & 6.2 & $-0.1$ \\
Best-of-8 & 1.4 & 2.6 & 4.3 & 6.9 & 3.4 & 0.1 & 0.8 & 0.4 & 0.6 & 0.3 & 0.8 & 1.4 & 2.8 & 3.3 & 1.0 & 0.1 & 0.9 & 0.4 & 0.5 & 0.2 \\
FK-8~\cite{singhal2025fk} & 0.9 & 2.2 & 3.1 & 9.7 & 3.2 & $-0.2$ & 0.3 & 0.8 & 4.0 & $-0.2$ & 0.6 & 1.0 & 2.3 & 6.2 & 1.4 & $-0.2$ & 0.3 & 0.8 & 2.1 & $-0.9$ \\
\midrule
\textbf{DRG (ours)} & 98.9 & 98.8 & 98.7 & 99.8 & 97.4 & $\mathbf{3.2}$ & $\mathbf{5.0}$ & $\mathbf{4.9}$ & $\mathbf{11.0}$ & $\mathbf{14.6}$ & $\mathbf{49.9}$ & $\mathbf{47.4}$ & $\mathbf{52.2}$ & $\mathbf{53.9}$ & 42.4 & $\mathbf{2.3}$ & $\mathbf{4.7}$ & $\mathbf{3.8}$ & $\mathbf{6.5}$ & $\mathbf{13.1}$ \\
\bottomrule
\end{tabular}
\end{table*}

\paragraph{Datasets and predictors.}
We use three datasets in our experiments (\cref{tab:settings}). OpenKBP~\cite{babier2021openkbp} and GDP-HMM~\cite{gao2025gdphmm} are from two public challenges. We additionally include an in-house dataset (IRB approved) to add diversity in tumor sites and techniques. For OpenKBP, we use public methods DoseDiff~\cite{zhang2024dosediff} and MD-Dose~\cite{fu2024mddose}. To add diversity to model architecture, we also implement DoseDiT based on DoseDiff, replacing the UNet denoiser with a diffusion-transformer~\cite{peebles2023scalable} denoiser. Because no public diffusion predictors exist for GDP-HMM or the local cohort, we train DoseDiff models on both datasets.

\paragraph{DoseSteer evaluation protocol.} 
We evaluate the retrospective correction setting in which an unsteered prediction deviates from a physician's expectation and structure-level targets communicate the intended dose.
For each held-out case, the clinical reference plan $x^\star$ provides a realized outcome with a known 3D ground-truth dose; treating $x^\star$ as the proxy for the intended outcome and extracting its structure-level statistics $y_k = q_k(x^\star)$ (the OAR mean dose or the PTV $D_{95}$) therefore provides a self-consistent protocol for evaluating full-volume 3D dose fidelity against a valid ground truth. 
Every steering method receives the \textit{exact same} scalar targets $y_k$, isolating how faithfully each algorithm propagates sparse structure-level intent into a coherent dose distribution. We evaluate both agreement with the supplied targets and overall dose prediction quality, and additionally report error outside the steered structures to assess fidelity in unconstrained regions (counterfactual target-perturbation stress tests are provided in Supp.).

\paragraph{Metrics.}
Following the official benchmark protocols of OpenKBP~\cite{babier2021openkbp} and GDP-HMM~\cite{gao2025gdphmm}, we evaluate overall dose quality using the challenge-specific dose error, the DVH score, and outside-structure error.
For a predicted dose $x$ and clinical dose $x^\star$ over body voxels $\mathcal{B}$, the OpenKBP dose and DVH scores are
\begin{equation}\label{eq:metrics}
\begin{gathered}
e_{\mathrm{dose}}=\frac{1}{|\mathcal{B}|}\sum_{v\in\mathcal{B}}|x_v-x^\star_v|,\\
e_{\mathrm{DVH}}=\frac{1}{|P|}\sum_{D\in P}|D(x)-D(x^\star)|,
\end{gathered}
\end{equation}
where $P$ collects the standard clinical DVH criteria ($D_1$, $D_{95}$, $D_{99}$ for PTVs; $D_{0.1\mathrm{cc}}$ and mean dose for OARs, plus target $D_{\max}$ on the local cohort).
On OpenKBP, the official metric is $e_{\text{dose}}$; on GDP-HMM, it follows the challenge definition as the mean voxel error over the union of voxels exceeding 5 Gy in $x$ or $x^\star$, normalized by the count of voxels exceeding 5 Gy in $x^\star$. For the local cohort, which exhibits lower peripheral brain doses, we apply the same thresholded metric at 2 Gy.
Outside error is computed analogously over body voxels outside the union of all steered structures (above 5 and 2\,Gy in $x$ or $x^\star$ on GDP-HMM and the local cohort), while target error is $\frac{1}{K}\sum_k \vert{}q_k(x) - y_k\vert{}$. All table entries report percentage error reduction relative to the unsteered prediction (absolute Gy values are in Supp.).

\paragraph{Baselines.}
We compare with twelve training-free methods given the same targets: a direct edit of the estimate, DDNM~\cite{wang2023ddnm}, DiffPIR~\cite{zhu2023diffpir}, $\Pi$GDM~\cite{song2023pigdm}, DPS~\cite{chung2023dps}, TMPD~\cite{boys2024tmpd}, TFG~\cite{ye2024tfg}, DAPS~\cite{zhang2025daps}, DDiff~\cite{kim2026ddiff}, DNO~\cite{tang2025dno}, best-of-8 and FK steering with 8 particles~\cite{singhal2025fk}.

\paragraph{Implementation.}
Following DoseDiff~\cite{zhang2024dosediff}, the unsteered predictors and DRG use deterministic DDIM over $T=8$ steps, with DRG steering the last four steps, $\mathcal{S}=\{3,2,1,0\}$.
Baselines follow their published algorithms with validation-tuned hyperparameters (details in Supp.).

\subsection{Main results}
\label{sec:main_results}

\paragraph{Target error.}
Meeting the targets is the first goal of steering.
DRG removes nearly all of the target error, $97.4$--$99.8\,\%$ on the five settings (\cref{tab:main}), more than every comparison method except the direct edit, DDNM and DiffPIR.
These three remove essentially all of it because they impose the targets on the estimate directly, shifting the dose inside each structure to its target: the direct edit on the final output, DDNM and DiffPIR on the clean estimate at every step.
Every other method removes at most $84\,\%$ (DNO on GDP-HMM), and gradient-based guidance $3$--$79\,\%$.

\begin{figure}[t]
\centering
\includegraphics[width=\linewidth]{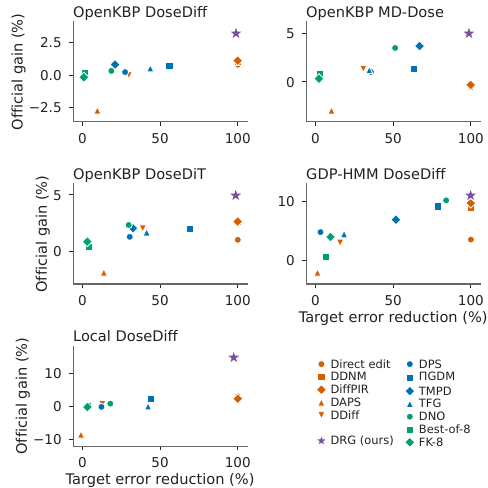}
\caption{\textbf{DRG can meet the targets while reducing overall dose error.}
Target error reduction against the gain in the official metric, both relative to the unsteered prediction (\cref{tab:main}); color: method category, marker: method.
The direct edit, DDNM and DiffPIR sit at $100\,\%$ with smaller official gains; every other method removes at most $84\,\%$ of the target error (DNO on GDP-HMM).}
\label{fig:tension}
\end{figure}

\begin{table}[t]
\centering
\caption{\textbf{Ablation results on key designs of DRG.}
Official metric gain (\%) on test sets when removing the price, the shape term, or both (top), or restricting DRG's structure-wise joint solver to a single lever (bottom).
On GDP-HMM the validated $\lambda$ is $0$, so there is no shape term to remove and the third row equals the first.}
\label{tab:ablation}
\footnotesize
\setlength{\tabcolsep}{5pt}
\begin{tabular}{lccccc}
\toprule
\textbf{Variant} & \textbf{DD} & \textbf{MD} & \textbf{DiT} & \textbf{GDP} & \textbf{Local} \\
\midrule
No price ($\mu\to0$) & 3.3 & 4.5 & 4.2 & 7.7 & 15.0 \\
No shape term ($\lambda=0$) & 2.9 & 4.8 & 4.5 & -- & 14.2 \\
No price, no shape term & 1.7 & 3.5 & 4.2 & 7.7 & 11.9 \\
\midrule
Noise lever only & 1.1 & 5.4 & 5.1 & 11.0 & 13.1 \\
Context lever only & 2.6 & $-0.1$ & 2.9 & 1.3 & 6.0 \\
DRG (full) & 3.2 & 5.0 & 4.9 & 11.0 & 14.6 \\
\bottomrule
\end{tabular}
\end{table}

\begin{figure}[t]
\centering
\includegraphics[width=\linewidth]{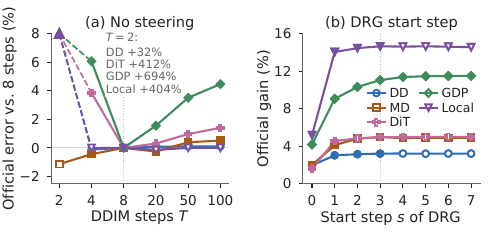}
\caption{\textbf{Ablation on DDIM steps and steering steps.}
(a)~Official error of the unsteered predictors against the number of DDIM steps, relative to 8 steps (positive: worse); two steps are off the scale on four settings (triangles, values printed).
(b)~Gain in the official metric when DRG corrects the steps $t=s,\dots,0$, for every start step $s$; hollow markers: not significantly different from $s=3$; the curves of MD-Dose and DoseDiT nearly coincide.
}
\label{fig:steps}
\vspace{-4pt}%
\end{figure}

\begin{figure}[t]
\centering
\includegraphics[width=\linewidth]{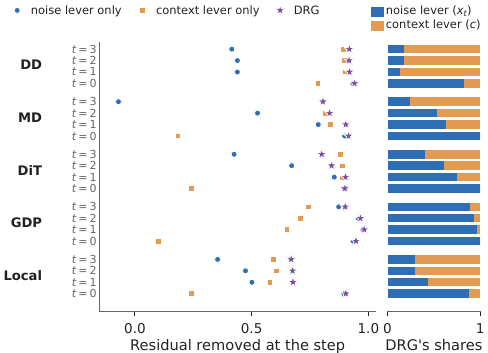}
\caption{\textbf{DRG leans on the lever that works.}
Points: share of the target residual removed at each steering step, $1-\sum_k|r_k|$ after the step's correction over $\sum_k|r_k|$ before, by the noise or context lever alone, and DRG, mean over the test plans. Bars: DRG's split of the linearized metric change, $\lVert G^N\alpha\rVert/(\lVert G^N\alpha\rVert+\lVert G^C\beta\rVert)$ for the noise lever, mean over the cases.}
\label{fig:shares}
\end{figure}

\paragraph{Overall dose prediction quality.}
DRG reduces the official dose error by $3.2$ to $14.6\,\%$ ($7.8\,\%$ on average) and outperforms every comparison method on all five settings, by $0.8$ to $11.6$ points; on the per-case official error it is better than every method on every setting (paired Wilcoxon, $p<0.05$).
It also has the best DVH score on four settings.
Moreover, although targets constrain only the steered structures, \textbf{DRG also lowers outside-structure error by 2.3\% to 13.1\%, outperforming all baselines on every setting}. By contrast, nine of the twelve baselines worsen outside error on the local cohort, and while direct edit, DDNM, and DiffPIR satisfy the targets to over 99.9\%, their smaller or negative official gains (such as $-0.3\%$ on MD-Dose) confirm that \textbf{matching scalar targets alone cannot recover the clinical dose distribution} (\cref{fig:tension}).

\subsection{Ablation studies}
\label{sec:ablations}

\begin{figure*}[t]
\centering
\includegraphics[width=\textwidth]{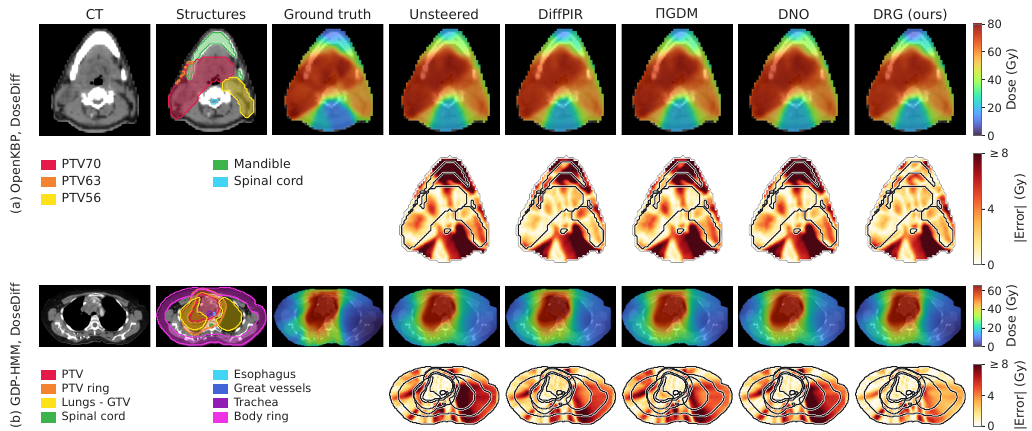}
\caption{\textbf{Qualitative results.}
Axial slices from (a) OpenKBP and (b) GDP-HMM test cases with DoseDiff, comparing unsteered predictions, the top baseline from each category (DiffPIR, $\Pi$GDM, DNO), and DRG (top: dose; bottom: absolute error saturated at 8 Gy). DRG achieves the lowest error both inside and outside the steered structures.
}
\label{fig:qual}
\end{figure*}

\paragraph{Price and shape term.}
Removing the shape penalty ($\lambda = 0$) lowers the official gain on every setting with a shape term, and removing the price ($\mu \to 0$) lowers it on MD-Dose, DoseDiT and GDP-HMM while changing it by at most 0.4 points on the other two; removing both causes the sharpest drop (for instance, from 14.6\% to 11.9\% on the local cohort and from 11.0\% to 7.7\% on GDP-HMM, \cref{tab:ablation}). 

\paragraph{Single lever vs. joint solve.}
Even with the noise lever alone, our structure-wise probing and joint solver outperform all 12 baselines (7.1\% vs. $<3.5\%$ mean gain), yet the two levers vary across predictors (\cref{tab:ablation}). Adding the context lever lifts the official gain on OpenKBP DoseDiff (1.1\% to 3.2\%) and Local (13.1\% to 14.6\%). Where the noise lever dominates, the joint solve matches it on GDP-HMM (11.0\%) and sits slightly below it on MD-Dose and DoseDiT (5.0\% vs. 5.4\% and 4.9\% vs. 5.1\%). With no per-predictor lever selection, the joint solve is never more than 0.4 points below the better single lever and averages 7.8\%, surpassing an oracle per-setting single-lever choice (7.4\%).

\paragraph{Sampling and steering steps.}
Eight DDIM steps are the best or within $1.1\,\%$ of the best step count for the unsteered predictors (\cref{fig:steps}a), and corrections from $t=3$ on are the knee (\cref{fig:steps}b): starting later lowers the gain on every setting, while starting earlier changes it by at most $0.5$ points.

\paragraph{Lever shares.}
 DRG shifts weight to whichever lever is effective at each step (\cref{fig:shares}). At $t = 3$ on MD-Dose and DoseDiT, and across $t \in \{3, 2, 1\}$ for DoseDiff on OpenKBP and Local, the noise lever alone removes at most half of the residual, so DRG assigns most of the correction (56\% to 86\%) to the context lever. On MD-Dose and DoseDiT, the noise lever strengthens at $t \in \{2, 1\}$ and takes the larger share (54\% to 76\%); at the final step ($t = 0$) and across all steps on GDP-HMM, where the noise lever already removes over 85\% of the residual, DRG relies primarily on the noise lever.

\subsection{Qualitative results}
\label{sec:qualitative}

\Cref{fig:qual} compares DRG against the strongest baseline from each category (DiffPIR, $\Pi$GDM, and DNO) on OpenKBP and GDP-HMM cases. On both slices, DRG achieves the lowest error both inside the steered structures (1.9 and 2.1 Gy vs. 3.4 and 3.6 Gy unsteered) and in unconstrained outside regions (4.3 and 2.6 Gy vs. 5.7 and 4.5 Gy unsteered, whereas baselines shift outside error by at most 0.4 Gy).

\section{Limitations \& Future Work}
\label{sec:limitations}

\paragraph{Compute overhead.}
Because DRG probes the denoiser per target structure at each steered step, it requires more network evaluations than direct projection methods such as DiffPIR (details in Supp.).
Developing more efficient steering algorithms is a promising direction in the future.

\paragraph{Diffusion backbone availability.}
Because diffusion-based dose prediction is still an emerging paradigm, DoseDiff and MD-Dose are currently the only open-sourced public diffusion predictors.
While we expanded architectural and cohort coverage by training DoseDiT and multi-site DoseDiff models across three datasets, applying DRG to stronger, next-generation diffusion predictors as they become available will further amplify its generalization ability.

\paragraph{Gap to complex clinical workflows.}
As an initial study on inference-time dose steering, we establish a standardized benchmark using well-defined equality targets.
Real-world planning often involves partial target specification, one-sided inequality constraints (\eg, upper bounds on OAR dose), and prioritized trade-offs among competing structures. Moreover, verifying that steered dose gains carry over to physically deliverable plans through downstream treatment planning system (TPS) optimization and clinician-in-the-loop studies remains important future work.

\section{Conclusion}
\label{sec:conclusion}
We introduced DoseSteer, a training-free framework that steers frozen diffusion dose predictors towards structure-level clinical targets at inference time.
Its core algorithm, Dual-Response Guidance, couples structure-wise noise and context probing with a regularized joint solver across all target structures, meeting clinical targets while preserving full-volume dose quality. Across five dataset-model configurations, DRG outperforms 12 training-free methods, removing at least $97\,\%$ of the target error while reducing the official dose error by 7.8\% on average relative to the unsteered predictions.
More broadly, DoseSteer bridges learned anatomical dose priors and explicit structure--level clinical targets at inference time, enabling target-guided dose refinement prior to plan optimization.

{
    \small
    \bibliographystyle{ieeenat_fullname}
    \bibliography{main}
}

\clearpage
\setcounter{page}{1}
\makeatletter
\let\supp@addcontentsline\addcontentsline
\def\supp@toc{toc}
\renewcommand*\addcontentsline[3]{\supp@addcontentsline{#1}{#2}{#3}\def\supp@ext{#1}%
  \ifx\supp@ext\supp@toc\supp@addcontentsline{stc}{#2}{#3}\fi}
\newcommand\suppcontents{\section*{Contents}%
  \begingroup\c@tocdepth=2 \renewcommand*\l@subsection{\@dottedtocline{2}{1.5em}{2.9em}}\@starttoc{stc}\endgroup}
\makeatother
\maketitlesupplementary
\suppcontents
\appendix

\section{Proofs}
\label{sec:proofs}

This section proves \cref{lem:tweedie} and \cref{prop:tilt,prop:responses,prop:surface} of \cref{sec:method}; it also states and proves \cref{lem:manifold}, the surface picture of \cref{sec:levers} under the manifold hypothesis, and \cref{prop:solve}, a property of the solve of \cref{sec:drg}.
Except for \cref{prop:solve}, these statements describe the data distribution and an ideal denoiser (A0): they motivate the design of DRG and do not show that the posterior interpretation holds exactly for the trained predictors.
DRG does not need this interpretation to hold: it uses $N_k$ and $C_k$ as measured responses of the trained network and weighs them in the solve by their measured effect on the target metrics; only the scale $v_k$ of the price is derived from (A0), and \cref{prop:solve} does not rely on it.

\paragraph{Notation.}
Fix $c$ and $t$, and write $\sigma_t^2=1-\bar\alpha_t$, so that $\kappa_t=\sigma_t^2/\sqrt{\bar\alpha_t}$.
The prior $p(\cdot\mid c)$ is a probability measure on $\mathbb{R}^n$, and $\varphi(u)=(2\pi\sigma_t^2)^{-n/2}\exp(-\lVert u\rVert^2/(2\sigma_t^2))$ is the density of the scaled noise $\sigma_t\varepsilon$.
$A\in\mathbb{R}^{K\times n}$ has the rows $A_j=\mathbf{1}_j^\top/|\mathbf{1}_j|$, so that $m_j(x)=A_jx$, and $\Delta_\vartheta=\sum_k\alpha_kN_k+\sum_k\beta_kC_k$ is the correction with strengths $\vartheta$.
In \cref{prop:responses,prop:surface}, $\delta=\sum_k\alpha_k\mathbf{1}_k$ is the push of the noise lever on $x_t$ and $a=\delta/\kappa_t$.
Norms are Euclidean.

\subsection{Tweedie's identities and the two re-weightings}
\label{sec:supp-tilt}
\begin{proof}[Proof of \cref{lem:tweedie}]
Both identities are classical~\cite{robbins1956empirical,efron2011tweedie,kadkhodaie2021stochastic,meng2021highorder,manor2024posterior}; we recall the derivation because \cref{prop:tilt,prop:responses,prop:surface} reuse its exponential-family form.
Expanding the square in the Gaussian density,
\begin{equation}\label{eq:supp-expfam}
\begin{gathered}
\varphi\big(x_t-\sqrt{\bar\alpha_t}\,x_0\big)=\varphi(x_t)\,e^{\eta^\top x_0}\,e^{-\bar\alpha_t\lVert x_0\rVert^2/(2\sigma_t^2)},\\
\eta=\frac{x_t}{\kappa_t},
\end{gathered}
\end{equation}
so with $\nu(dx_0)=\exp(-\bar\alpha_t\lVert x_0\rVert^2/(2\sigma_t^2))\,p(dx_0\mid c)$ and $\Psi(\eta)=\log\int\exp(\eta^\top x_0)\,\nu(dx_0)$, Bayes' rule reads
\begin{equation}\label{eq:supp-posterior}
p(dx_0\mid x_t,c)=e^{\eta^\top x_0-\Psi(\eta)}\,\nu(dx_0).
\end{equation}
The posterior is an exponential family in $\eta$, and the derivatives of its log-partition function $\Psi$ are its cumulants ($\Psi$ is finite, hence smooth, because the Gaussian factor in $\nu$ dominates $e^{\eta^\top x_0}$ for every $\eta$); with $\tilde x$ the deviation of $x_0$ from its mean under the law at hand, here the posterior,
\begin{equation}\label{eq:supp-cumulants}
\begin{gathered}
\nabla\Psi(\eta)=\mathbb{E}[x_0\mid x_t,c],\qquad
\nabla^2\Psi(\eta)=\operatorname{Cov}[x_0\mid x_t,c],\\
\nabla^3\Psi(\eta)[u,u,w]=\mathbb{E}\big[(u^\top\tilde x)^2\,w^\top\tilde x\mid x_t,c\big].
\end{gathered}
\end{equation}

\emph{Mean.}
Taking the conditional expectation of $x_0=(x_t-\sigma_t\varepsilon)/\sqrt{\bar\alpha_t}$ gives the first identity with $\varepsilon^\star=\mathbb{E}[\varepsilon\mid x_t,c]$, the $L^2$ projection of $\varepsilon$ on the functions of $(x_t,c)$ and hence the minimizer of the noise-prediction loss; (A0) states $\epsilon_\theta=\varepsilon^\star$, \ie $f_\theta=\mathbb{E}[x_0\mid x_t,c]$.

\emph{Covariance.}
Since $\eta=x_t/\kappa_t$, the chain rule and \cref{eq:supp-cumulants} give $\partial\,\mathbb{E}[x_0\mid x_t,c]/\partial x_t=\nabla^2\Psi(\eta)/\kappa_t=\operatorname{Cov}[x_0\mid x_t,c]/\kappa_t$, the second identity; under (A0) the two sides are $J$ and $\Sigma_t/\kappa_t$, so $\Sigma_t=\kappa_tJ$ and $J$ is symmetric positive semi-definite.
\end{proof}

\begin{proof}[Proof of \cref{prop:tilt}]
\emph{Push.}
By \cref{eq:supp-posterior}, moving $x_t$ to $x_t+\delta$ moves the natural parameter from $\eta$ to $\eta+a$ with $a=\delta/\kappa_t$, so, for any prior,
\begin{equation}\label{eq:supp-tilt}
\begin{gathered}
p(dx_0\mid x_t+\delta,c)=p(dx_0\mid x_t,c)\,e^{a^\top x_0-\Lambda(a)},\\
\Lambda(a)=\Psi(\eta+a)-\Psi(\eta)=\log\mathbb{E}\big[e^{a^\top x_0}\mid x_t,c\big],
\end{gathered}
\end{equation}
the first identity; $\Lambda$ is the cumulant generating function of the posterior at $x_t$.

\emph{Context.}
The noise does not depend on $c$, so $p(x_t\mid x_0,c)=p(x_t\mid x_0)$, and Bayes' rule for $c=(c_{\setminus k},c_k)$ gives
\begin{equation*}
\begin{aligned}
p(x_0\mid x_t,c)&\propto p(x_0\mid c_{\setminus k})\,p(c_k\mid x_0,c_{\setminus k})\,p(x_t\mid x_0)\\
&\propto p(x_0\mid x_t,c_{\setminus k})\,e^{\ell_k(x_0)},
\end{aligned}
\end{equation*}
the second identity.
Neither identity involves the network: both are statements about the joint law of $(x_0,c)$ and the Gaussian noising.
In the second, $c_{\setminus k}$ stands for the context without structure $k$; DRG evaluates the network with the channels of structure $k$ set to zero (\cref{sec:supp-solve}), which we use as an approximation of this context, and the same form holds exactly with this input as the reference, as shown next.
\end{proof}

\paragraph{The zeroed input as the reference.}
Write $c^0_k$ for this zeroed input, which \cref{sec:prelim} writes $c_{\setminus k}$ and which presents structure $k$ as absent.
Applying the second identity to $c$ and to $c^0_k$, which differ only in these channels, and dividing gives, if structure $k$ can be absent ($p(0\mid x_0,c_{\setminus k})>0$ for $p(\cdot\mid x_t,c)$-almost every $x_0$),
\begin{equation*}
\begin{gathered}
p(x_0\mid x_t,c)\propto p(x_0\mid x_t,c^0_k)\,e^{\ell_k(x_0)-\ell^0_k(x_0)},\\
\ell^0_k(x_0)=\log p(0\mid x_0,c_{\setminus k}),
\end{gathered}
\end{equation*}
with $\ell^0_k$ the log-likelihood that structure $k$ is absent: the zeroed input is the reference, and the log-likelihood ratio $\ell_k-\ell^0_k$ of the contour against its absence is the weight.
\Cref{prop:responses,prop:surface} and their proofs use only this form, so they hold with $c^0_k$ in place of $c_{\setminus k}$ and $\ell_k-\ell^0_k$ in place of $\ell_k$, under the integrability and moment assumptions of \cref{sec:supp-responses}, the integrability being automatic for $0\le w\le1$.
(A0) at $c_{\setminus k}$ then becomes (A0) at $c^0_k$: the output at the zeroed input is the posterior mean given that structure $k$ is absent, the minimizer of the denoising loss over plans without structure $k$.
Under (A0), the interpretation thus needs no marginalization over the contour, only training plans that lack structure $k$; where none does, $c^0_k$ is an extrapolation (\cref{sec:supp-solve}).

\subsection{The two responses}
\label{sec:supp-responses}
\paragraph{Remainders of \cref{prop:responses}.}
On every metric, the proof below bounds $R$ by a constant times the largest posterior variance of $a^\top x_0$ along the push, which at $x_t$ is at most $K$ times the noise part of the price $\mathcal{P}$, and $R_k$ by a constant times $|\beta_k|(1+|\beta_k|)$ times the largest variance of $\ell_k(x_0)$ under $q_w$ for $w$ in the interval spanned by $0$, $1$ and $1+\beta_k$.
For the context, the network is evaluated only at the two ends of the re-weighting, $w=0$ and $w=1$, where under (A0) it returns $\bar{x}(0)$ and $\bar{x}(1)$, as in classifier-free guidance, so DRG extends the chord between them, the mean slope of $\bar{x}$ on $[0,1]$, rather than the tangent at $w=1$; unlike $R$, $R_k$ therefore has a part linear in $\beta_k$, the gap between chord and tangent. Both parts vanish with the variance of $\ell_k(x_0)$ under the laws $q_w$, and the extension is typically short: the median $|\beta_k|$ is below $0.2$ on every setting.

\begin{proof}[Proof of \cref{prop:responses}]
Under (A0) the network returns posterior means, so $f_\theta(x_t+\delta,c)$ is the mean of the pushed posterior of \cref{prop:tilt}, and $f_\theta(x_t,c_{\setminus k})$ and $f_\theta(x_t,c)$ are the means of the posteriors under $c_{\setminus k}$ and $c$, the former approximately, and exactly with the zeroed input $c^0_k$ as the reference (\cref{sec:supp-tilt}).

\emph{(i), Gaussian posterior.}
If the posterior at $x_t$ is $\mathcal{N}(\bar{x},\bar\Sigma)$, then $\Lambda(a)=a^\top\bar{x}+\frac12a^\top\bar\Sigma a$ and the re-weighted law is $\mathcal{N}(\bar{x}+\bar\Sigma a,\bar\Sigma)$; under (A0), $\bar{x}=\hat{x}_0$ and $\bar\Sigma=\kappa_tJ$ by \cref{lem:tweedie}, so $f_\theta(x_t+\delta,c)=\hat{x}_0+J\delta$ exactly.

\emph{(i), general posterior.}
We bound the remainder $R(\delta)=f_\theta(x_t+\delta,c)-\hat{x}_0-J\delta$.
Let $\zeta(s)=\mathbb{E}[x_0\mid x_t+s\delta,c]=\nabla\Lambda(sa)$ for $s\in[0,1]$; then $\zeta(0)=\hat{x}_0$, $\zeta'(0)=\Sigma_ta=J\delta$ and $\zeta(1)=f_\theta(x_t+\delta,c)$, so Taylor's formula with integral remainder gives
\postdisplaypenalty=10000
\begin{equation}\label{eq:supp-remainder}
R(\delta)=\int_0^1(1-s)\,\nabla^3\Lambda(sa)[a,a,\cdot\,]\,ds .
\end{equation}
\postdisplaypenalty=0
By \cref{eq:supp-cumulants} at $\eta+sa$, the $j$-th metric of the integrand is a third central moment of the posterior at $x_t+s\delta$,
\begin{equation*}
\nabla^3\Lambda(sa)[a,a,A_j^\top]=\mathbb{E}_s\big[(a^\top\tilde x)^2\,A_j\tilde x\big],
\end{equation*}
which vanishes for a Gaussian.
Every dose value lies in an interval of width $W$, so does every metric, hence $|A_j\tilde x|\le W$, the integrand is at most $W\operatorname{Var}[a^\top x_0\mid x_t+s\delta,c]$, and since $1-s$ integrates to $1/2$ over $s\in[0,1]$,
\begin{equation*}
|A_jR(\delta)|\le\frac{W}{2}\sup_{s}\operatorname{Var}[a^\top x_0\mid x_t+s\delta,c],
\end{equation*}
the constant behind \cref{prop:responses}(i); since $\operatorname{Var}[a^\top x_0]\le nW^2\lVert a\rVert^2$, the same steps voxel by voxel give $R(\delta)=O(\lVert\delta\rVert^2)$.

\emph{(i), the price.}
We show that this variance, at $x_t$ ($s=0$), is at most $K$ times the noise part of the price.
Along the push it changes by the third cumulant of $a^\top x_0$, of third order in the push, so the price controls the remainder to leading order only.
For the push $\delta=\sum_k\alpha_k\mathbf{1}_k$, $a^\top x_0=\kappa_t^{-1}\sum_k\alpha_k|\mathbf{1}_k|\,m_k(x_0)$, and with $N_k=J\mathbf{1}_k$ and \cref{lem:tweedie},
\begin{equation*}
\operatorname{Var}[m_k(x_0)\mid x_t,c]=\frac{\mathbf{1}_k^\top\Sigma_t\mathbf{1}_k}{|\mathbf{1}_k|^2}=\frac{\kappa_tG^N_{kk}}{|\mathbf{1}_k|}=v_k ,
\end{equation*}
which is \cref{eq:vk}; hence $|\mathbf{1}_k|/\kappa_t=G^N_{kk}/v_k$ for $v_k>0$, while $v_k=0$ gives $\Sigma_t\mathbf{1}_k=0$ and structure $k$ adds nothing to the variance below.
Minkowski's inequality for standard deviations and then the Cauchy--Schwarz inequality give
\begin{align*}
\operatorname{Var}[a^\top x_0\mid x_t,c]^{1/2}&\le\sum_k\frac{|\alpha_kG^N_{kk}|}{v_k^{1/2}}\\
&\le\Big(K\sum_k\frac{\lVert G^N_{:,k}\alpha_k\rVert^2}{v_k}\Big)^{1/2};
\end{align*}
for a single pushed structure the variance is $(G^N_{kk}\alpha_k)^2/v_k$, the diagonal part of the price.

\emph{(i), price and divergence.}
Finally, the same variance measures how far the push moves the belief.
Let $\mathrm{KL}_\delta$ be the Kullback--Leibler divergence of the pushed posterior $p(\cdot\mid x_t+\delta,c)$ from $p(\cdot\mid x_t,c)$; by \cref{eq:supp-tilt},
\begin{align*}
\mathrm{KL}_\delta&=a^\top\nabla\Lambda(a)-\Lambda(a)\\
&=\int_0^1s\operatorname{Var}[a^\top x_0\mid x_t+s\delta,c]\,ds\\
&=\tfrac12\operatorname{Var}[a^\top x_0\mid x_t,c]+O(\lVert a\rVert^3),
\end{align*}
so, to second order, the variance that bounds the remainder is twice the divergence the push causes.
One consequence supports \cref{sec:levers}: the Cauchy--Schwarz inequality in the inner product of $\Sigma_t$ gives $m_k(J\delta)^2\le v_k\operatorname{Var}[a^\top x_0\mid x_t,c]$ for every push, with equality when $\delta$ is a multiple of $\mathbf{1}_k$, so a push that changes $m_k$ by $\tau$ to first order causes a divergence of at least $\tau^2/(2v_k)$ to second order, attained by the noise lever's own push, and no push moves $m_k$ to first order when $v_k=0$.

\emph{(ii), the family $q_w$.}
By \cref{prop:tilt}, $q_0$ and $q_1$ are the posteriors under $c_{\setminus k}$ and $c$, and $q_w$ is an exponential family in $w$ with sufficient statistic $\ell_k$.
For a discrete $c_k$ such as a mask, $\ell_k\le0$ and $q_w$ is defined for every $w\ge0$; in all cases we assume $e^{w\ell_k}$ integrable and the moments below finite for $w$ in the interval spanned by $0$, $1$ and $1+\beta_k$.

\emph{(ii), chord and curvature.}
We express $D_k$ through the family $q_w$.
Write $\tilde\ell=\ell_k(x_0)-\mathbb{E}_{q_w}\ell_k$.
Differentiating the log-partition function as in \cref{eq:supp-cumulants},
\begin{equation*}
\bar{x}'(w)=\mathbb{E}_{q_w}[\tilde x\,\tilde\ell],\qquad \bar{x}''(w)=\mathbb{E}_{q_w}[\tilde x\,\tilde\ell^{\,2}].
\end{equation*}
Under (A0) at $c_{\setminus k}$ and $c$, $\bar{x}(0)=f_\theta(x_t,c_{\setminus k})$ and $\bar{x}(1)=f_\theta(x_t,c)=\hat{x}_0$, so $D_k=\bar{x}(1)-\bar{x}(0)$, the integral of $\bar{x}'(w)$ over $w\in[0,1]$: the chord is the mean slope of $\bar{x}$ on $[0,1]$.

\emph{(ii), extension.}
We bound the remainder of the linear extension, $R_k(\beta)=\bar{x}(1+\beta)-\hat{x}_0-\beta D_k$ with $\beta=\beta_k$: it is the gap between the chord and the tangent at $w=1$ plus the Taylor remainder beyond it,
\begin{equation*}
\begin{gathered}
R_k(\beta)=\beta\big[\bar{x}'(1)-D_k\big]\\
\qquad+\big[\bar{x}(1+\beta)-\bar{x}(1)-\beta\bar{x}'(1)\big],\\
\bar{x}'(1)-D_k=\textstyle\int_0^1\!\int_u^1\bar{x}''(v)\,dv\,du,\\
\bar{x}(1+\beta)-\bar{x}(1)-\beta\bar{x}'(1)\\
\qquad=\beta^2\textstyle\int_0^1(1-s)\,\bar{x}''(1+s\beta)\,ds .
\end{gathered}
\end{equation*}
Hence $|A_jR_k(\beta)|\le\big((|\beta|+\beta^2)/2\big)\sup_w|A_j\bar{x}''(w)|$, the supremum over the interval of $w$ spanned by $0$, $1$ and $1+\beta$.
Since $|A_j\tilde x|\le W$, $|A_j\bar{x}''(w)|\le W\operatorname{Var}_{q_w}[\ell_k(x_0)]$, so $|A_jR_k(\beta)|\le\big(|\beta|(1+|\beta|)/2\big)W\sup_w\operatorname{Var}_{q_w}[\ell_k(x_0)]$, the constant behind \cref{prop:responses}(ii); both brackets vanish when $\bar{x}''\equiv0$, \ie when $\bar{x}(w)$ is affine in $w$.
\end{proof}

\subsection{The surface of plausible doses}
\label{sec:supp-surface}
The manifold hypothesis of \cref{sec:levers}, as used in guided diffusion~\cite{chung2022mcg,he2024mpgd}, states that $p(x_0\mid c)$ is supported on a low-dimensional set $\mathcal{M}_c\subset\mathbb{R}^n$.
We call a dose plausible under $c$ if it lies on or near $\mathcal{M}_c$ (\cref{lem:manifold} bounds the distance of the estimates); here and in the main text, the word is used in this sense only, relative to the data distribution that the predictor models.
The posterior is absolutely continuous with respect to the prior and is therefore also supported on $\mathcal{M}_c$.

\begin{lemma}[Estimates and the surface]\label{lem:manifold}
Assume (A0) and the manifold hypothesis.
(i)~$\operatorname{dist}(f_\theta(x_t,c),\mathcal{M}_c)\le\sqrt{\operatorname{tr}\Sigma_t}$ for every $x_t$.
(ii)~If $\mathcal{M}_c$ lies in an affine subspace $\mathcal{H}_c$ with direction space $\mathcal{T}_c$, then $f_\theta(x_t,c)\in\mathcal{H}_c$ and $\operatorname{range}J\subseteq\mathcal{T}_c$ for every $x_t$; if $\mathcal{M}_c$ is moreover closed and convex, then $f_\theta(x_t,c)\in\mathcal{M}_c$ for every $x_t$, so every push of $x_t$ keeps $\hat{x}_0$ on $\mathcal{M}_c$.
In the setting of (ii), a change $u$ of $\hat{x}_0$ with a component $u_\perp\neq0$ orthogonal to $\mathcal{T}_c$ takes $\hat{x}_0$ to a distance at least $\lVert u_\perp\rVert$ from $\mathcal{M}_c$ and is reproduced by no push of $x_t$.
\end{lemma}
\begin{proof}
(i)~Under the posterior, $x_0\in\mathcal{M}_c$ almost surely, so $\operatorname{dist}(f_\theta(x_t,c),\mathcal{M}_c)\le\lVert x_0-f_\theta(x_t,c)\rVert$ almost surely; taking the expectation and applying Jensen's inequality give, since $f_\theta(x_t,c)$ is the posterior mean,
\begin{align*}
\operatorname{dist}(f_\theta(x_t,c),\mathcal{M}_c)&\le\mathbb{E}\big[\lVert x_0-f_\theta(x_t,c)\rVert\,\big|\,x_t,c\big]\\
&\le\mathbb{E}\big[\lVert x_0-f_\theta(x_t,c)\rVert^2\,\big|\,x_t,c\big]^{1/2}\\
&=\sqrt{\operatorname{tr}\Sigma_t}.
\end{align*}
Since the conditional mean minimizes the mean squared error, $\mathbb{E}\operatorname{tr}\Sigma_t\le\mathbb{E}\lVert x_0-x_t/\sqrt{\bar\alpha_t}\rVert^2=n(1-\bar\alpha_t)/\bar\alpha_t$, so on average the bound in (i) vanishes as $\bar\alpha_t\to1$.

(ii)~For $g\perp\mathcal{T}_c$, $g^\top x_0$ is constant on $\mathcal{H}_c$, hence almost surely under the posterior: the posterior mean lies in $\mathcal{H}_c$ and $g^\top\Sigma_tg=0$, so $\Sigma_tg=0$ and $\operatorname{range}J=\operatorname{range}\Sigma_t\subseteq\mathcal{T}_c$ by \cref{lem:tweedie}.
If $\mathcal{M}_c$ is closed and convex and $f_\theta(x_t,c)\notin\mathcal{M}_c$, a separating $g$ would give $g^\top f_\theta(x_t,c)>\sup_{x\in\mathcal{M}_c}g^\top x$, contradicting $g^\top f_\theta(x_t,c)=\mathbb{E}[g^\top x_0\mid x_t,c]$.
Every point of $\mathcal{M}_c$ and every $f_\theta(x_t+\delta,c)$ lie in $\mathcal{H}_c$, as does $\hat{x}_0$, so their difference from $\hat{x}_0+u$ has the component $-u_\perp$ orthogonal to $\mathcal{T}_c$ and norm at least $\lVert u_\perp\rVert$.
\end{proof}

\begin{proof}[Proof of \cref{prop:surface}]
We write $\hat{x}_0+\Delta-\bar{x}$ as five terms and bound each, under the assumptions of \cref{sec:supp-responses} for the joint tilt.

\emph{One family.}
By \cref{prop:tilt} the push and the $K$ switches tilt the same posterior.
With $\Lambda(a,\beta)=\log\mathbb{E}[\exp(a^\top x_0+\sum_k\beta_k\ell_k(x_0))\mid x_t,c]$, the mean of $q$ is
\begin{equation*}
\bar{x}(a,\beta)=\nabla_a\Lambda(a,\beta),
\end{equation*}
with $\bar{x}(0,0)=\hat{x}_0$, $\bar{x}(a,0)=\mathbb{E}[x_0\mid x_t+\delta,c]$ by \cref{eq:supp-tilt}, and $\bar{x}(0,\beta_k\mathbf{e}_k)$ the mean of $q_{1+\beta_k}$ in \cref{prop:responses}, $\mathbf{e}_k$ the $k$-th unit vector.

\emph{Decomposition.}
Using $J\delta=\sum_k\alpha_kJ\mathbf{1}_k$ and $C_k=D_k-(1-M_k)\odot D_k$,
\begin{equation*}
\hat{x}_0+\Delta-\bar{x}(a,\beta)=-T_1+T_2-T_3-T_4-T_5,
\end{equation*}
with
\begin{align*}
T_1&=\bar{x}(a,0)-\hat{x}_0-J\delta,\\
T_2&=\textstyle\sum_k\alpha_k\,(N_k-J\mathbf{1}_k),\\
T_3&=\textstyle\sum_k\big[\bar{x}(0,\beta_k\mathbf{e}_k)-\hat{x}_0-\beta_kD_k\big],\\
T_4&=\textstyle\sum_k\beta_k\,(1-M_k)\odot D_k,\\
T_5&=\bar{x}(a,\beta)-\bar{x}(a,0)-\textstyle\sum_k\big[\bar{x}(0,\beta_k\mathbf{e}_k)-\hat{x}_0\big].
\end{align*}
With $W$ as in the proof of \cref{prop:responses}, $\lVert\tilde x\rVert\le\Gamma:=\sqrt{n}\,W$ under every law below.
$T_1=R(\delta)$ is the remainder of \cref{prop:responses}(i), and \cref{eq:supp-remainder} with $\lVert\tilde x\rVert\le\Gamma$ gives $\lVert T_1\rVert\le(\Gamma/2)\sup\operatorname{Var}[a^\top x_0]$ along the push.
$\lVert T_2\rVert\le\sum_k|\alpha_k|e_k$ by the definition of $e_k$; for a denoiser that is $C^3$ along $\mathbf{1}_k$ the second-order Taylor terms of the central difference cancel and $e_k=O(h^2)$.
$\lVert T_3\rVert\le\sum_k|\beta_k|\rho_k$ with $\rho_k=\big((1+|\beta_k|)/2\big)\Gamma\sup_w\operatorname{Var}_{q_w}[\ell_k(x_0)]$, the proof of \cref{prop:responses}(ii) with $\lVert\tilde x\rVert\le\Gamma$ in place of $|A_j\tilde x|\le W$.
$\lVert T_4\rVert\le\sum_k|\beta_k|\lVert(1-M_k)\odot D_k\rVert$, the part of $D_k$ that $M_k$ leaves out.

\emph{Interaction.}
$T_5$ is the non-additivity of the tilts,
\begin{align*}
T_5&=\big[\bar{x}(a,\beta)-\bar{x}(a,0)-\bar{x}(0,\beta)+\hat{x}_0\big]\\
&\quad+\big[\bar{x}(0,\beta)-\hat{x}_0-\textstyle\sum_k\big(\bar{x}(0,\beta_k\mathbf{e}_k)-\hat{x}_0\big)\big];
\end{align*}
the first bracket is $\int_0^1\!\!\int_0^1\partial_a\partial_\beta\bar{x}(sa,t\beta)[a,\beta]\,ds\,dt$, the second a sum over pairs $k<l$ of such integrals of $\beta_k\beta_l\,\partial_{\beta_k}\partial_{\beta_l}\bar{x}$.
These mixed derivatives are third joint cumulants, $\partial_{\beta_k}\partial_{\beta_l}\bar{x}=\mathbb{E}[\tilde x\,\tilde\ell_k\tilde\ell_l]$ and $\partial_{\beta_k}\nabla_a\bar{x}\,[u]=\mathbb{E}[\tilde x\,(u^\top\tilde x)\,\tilde\ell_k]$ under the tilted law, each of norm at most $\Gamma$ times the product of the two standard deviations by the Cauchy--Schwarz inequality.
With $s_a$ and $s_k$ the largest standard deviations of $a^\top x_0$ and $\ell_k(x_0)$ under the tilted laws between the posterior and $q$,
\begin{equation*}
\lVert T_5\rVert\le E(\delta,\beta):=\Gamma\Big(s_a\sum_k|\beta_k|s_k+\tfrac12\big(\textstyle\sum_k|\beta_k|s_k\big)^2\Big),
\end{equation*}
second order in the two pushes.
Adding the five bounds proves the proposition.
The implementation clips the estimate to the dose range; clipping is the projection onto a box, hence 1-Lipschitz, and $\bar{x}$ lies in the range, so the bound also holds for the clipped estimate.
\end{proof}
Under the manifold hypothesis, the proof of \cref{lem:manifold}(i) applies to $q$, which is absolutely continuous with respect to the posterior, so $\operatorname{dist}(\hat{x}_0+\Delta,\mathcal{M}_c)\le\lVert\hat{x}_0+\Delta-\bar{x}\rVert+(\operatorname{tr}\operatorname{Cov}_q)^{1/2}$, whose first term \cref{prop:surface} bounds; this is the sense in which \cref{sec:drg} calls the corrected estimate plausible.

\subsection{The solve}
\label{sec:supp-solveprop}
The solve of \cref{sec:drg} has the following property; it does not rely on (A0) and needs positive prior variances, which $\sigma_\beta>0$ and the floor of \cref{sec:supp-solve} ensure.
\begin{proposition}[The linearized residual does not increase]\label{prop:solve}
For $\mu,B>0$, $\lambda\ge0$ and finite positive prior variances of all strengths, with $\mathcal{P}$ computed from them, $\Phi$ has a unique minimizer $\vartheta^\star$ over $|\beta_k|\le B$, and
\begin{equation*}
\lVert r-G\vartheta^\star\rVert^2\le\lVert r\rVert^2-\mu\,\mathcal{P}(\vartheta^\star)-\lambda\,\mathcal{V}(\vartheta^\star).
\end{equation*}
\end{proposition}
\noindent The positivity condition matters for a trained network: there the finite difference can give $G^N_{kk}\le0$, and \cref{eq:price} with $v_k\le0$ is undefined ($v_k=0$) or indefinite ($v_k<0$).

\begin{proof}[Proof of \cref{prop:solve}]
\emph{Existence and uniqueness.}
Because $\Delta_\vartheta$ is linear in $\vartheta$, $\mathcal{V}(\vartheta)=\vartheta^\top Q_{\mathcal{V}}\vartheta$ with $Q_{\mathcal{V}}\succeq0$; and $\mathcal{P}(\vartheta)=\vartheta^\top Q_{\mathcal{P}}\vartheta$ with $Q_{\mathcal{P}}$ diagonal and positive, its entries being the inverse prior variances.
Hence $\Phi(\vartheta)=\vartheta^\top H\vartheta-2r^\top G\vartheta+\lVert r\rVert^2$, with $H=G^\top G+\mu Q_{\mathcal{P}}+\lambda Q_{\mathcal{V}}\succ0$, is strictly convex and coercive, and the box $\{\vartheta:|\beta_k|\le B\}$ is closed, convex and non-empty: the minimizer $\vartheta^\star$ exists and is unique.

\emph{Residual.}
$\vartheta=0$ is feasible with $\Phi(0)=\lVert r\rVert^2$, so $\lVert G\vartheta^\star-r\rVert^2+\mu\mathcal{P}(\vartheta^\star)+\lambda\mathcal{V}(\vartheta^\star)\le\lVert r\rVert^2$.
\end{proof}

\paragraph{Meeting the targets.}
If some $\bar\vartheta$ in the box meets the linearized targets, $G\bar\vartheta=r$, then $\lVert G\vartheta^\star-r\rVert^2\le\Phi(\vartheta^\star)\le\Phi(\bar\vartheta)=\mu\mathcal{P}(\bar\vartheta)+\lambda\mathcal{V}(\bar\vartheta)$; if $G^N$ is invertible, $\bar\vartheta=((G^N)^{-1}r,0)$ is such a point. A low-cost exact solution thus implies a small total linearized residual.

\paragraph{Exact residuals.}
$G$ is built from the same $N_k$ and $C_k$ as $\Delta_\vartheta$, so $m(\Delta_\vartheta)=\sum_k\alpha_kAN_k+\sum_k\beta_kAC_k=G\vartheta$ for every $\vartheta$, by construction; for an OAR, $y_k-m_k(\hat{x}_0+\Delta_\vartheta)=r_k-(G\vartheta)_k$.
For a PTV, $q_k$ is the 5th percentile of the dose over $\Omega_k$: it is monotone and shift-equivariant, $q_k(x+\tau\mathbf{1}_{\Omega_k})=q_k(x)+\tau$.
On $\Omega_k$ write $\Delta=\bar\Delta+\xi$ with $\bar\Delta=\operatorname{mean}_{\Omega_k}\Delta$.
Monotonicity gives $|q_k(\hat{x}_0+\Delta)-q_k(\hat{x}_0)-\bar\Delta|\le\max_{\Omega_k}|\xi|$, and the linear row, an average over the cold set $\mathbf{1}_k\subseteq\Omega_k$, satisfies $|A_k\Delta-\bar\Delta|\le\max_{\Omega_k}|\xi|$.
Hence the exact residual $y_k-q_k(\hat{x}_0+\Delta)$ and the linear one $r_k-(G\vartheta)_k$ differ by at most $2\max_{\Omega_k}|\Delta-\bar\Delta|$, which is zero for a $\Delta$ uniform on $\Omega_k$.

\paragraph{Clipping.}
The implementation clips $\hat{x}_0+\Delta$ to the dose range (\cref{sec:supp-models}); since $\hat{x}_0$ lies in the range, clipping moves every voxel to a value between its values in $\hat{x}_0$ and in $\hat{x}_0+\Delta$: it shrinks the correction voxel by voxel.
Clipping can still change a PTV's $D_{95}$ and an OAR's mean, so the residual of the clipped estimate is not covered by the proposition, and neither are the denoising steps that follow.

\section{Implementation details}
\label{sec:impl}

\subsection{Common setup}
\label{sec:supp-models}
We steer five trained predictors, each described below.
All five are trained from scratch with the noise-prediction loss of DDPM~\cite{ho2020ddpm} ($1000$ steps, variances rising linearly from $10^{-4}$ to $0.02$).
They are sampled with deterministic DDIM at the training timesteps $875,750,\dots,125,0$ ($t=7,\dots,0$).
After the network pass and again after the correction of \cref{alg:drg}, $\hat{x}_0$ is clipped to the dose range of the predictor; as in the unsteered sampler, the final dose is set to zero outside the possible body mask. All experiments run on a server with 8 NVIDIA RTX A6000 GPUs.

\subsection{DoseDiff on OpenKBP}
\label{sec:supp-dd}
\paragraph{Model source.}
Because no public model checkpoint is released for DoseDiff~\cite{zhang2024dosediff}, we train the model following the authors' public implementation.
Its denoiser is a UNet that sums the features of separate encoders for the noisy dose, the CT and the structure channels at every level and fuses them with a vision transformer at the bottleneck.

\paragraph{Inputs.}
The CT and 20 structure channels on axial $128\times128$ slices: a fused PTV map (PTV70, PTV63 and PTV56 weighted by $1$, $0.9$ and $0.8$), masks of the 7 OARs and the body, and signed distance maps of these 8 structures and the 3 PTVs.
CT and dose ($[0,80]$\,Gy) are mapped to $[-1,1]$.
For $c_{\setminus k}$, the mask and the distance map of structure $k$ are set to zero, and for a PTV also its term in the fused map.

\paragraph{Training.}
We train on the 200 training plans for 1600 epochs with AdamW (learning rate $2\times10^{-4}$, divided by 10 after 70\,\% of the epochs), batch size 64 and the authors' augmentation, and keep the checkpoint with the lowest validation MAE (the same for all other models).

\subsection{MD-Dose on OpenKBP}
\label{sec:supp-md}
\paragraph{Model source.}
MD-Dose~\cite{fu2024mddose} was developed on a private thoracic dataset, not on OpenKBP; because few diffusion dose predictors have public code (\cref{sec:limitations}), we adapt the authors' public implementation to OpenKBP.
Its denoiser is a UNet of Vision-Mamba blocks with parallel encoders for the noisy dose and the structures.
Only the data layer changes, an OpenKBP loader and a 10-channel instead of a 5-channel structure input; the diffusion process, the network and the training loop are the authors'.

\paragraph{Inputs.}
The CT, the fused PTV map and the masks of the 7 OARs and the body (no distance maps), on the slices of \cref{sec:supp-dd}; CT and dose ($[0,80]$\,Gy) are mapped to $[0,1]$.
For $c_{\setminus k}$, the mask of an OAR is set to zero, and a PTV's term is removed from the fused map.

\paragraph{Training.}
We train on the 200 training plans for 1600 epochs with the authors' settings (Adam at a constant learning rate of $10^{-4}$, batch size 26, no augmentation).

\subsection{DoseDiT on OpenKBP}
\label{sec:supp-dit}
\begin{figure}[t]
\centering
\includegraphics[width=\linewidth,trim=518 117 503 133.5,clip]{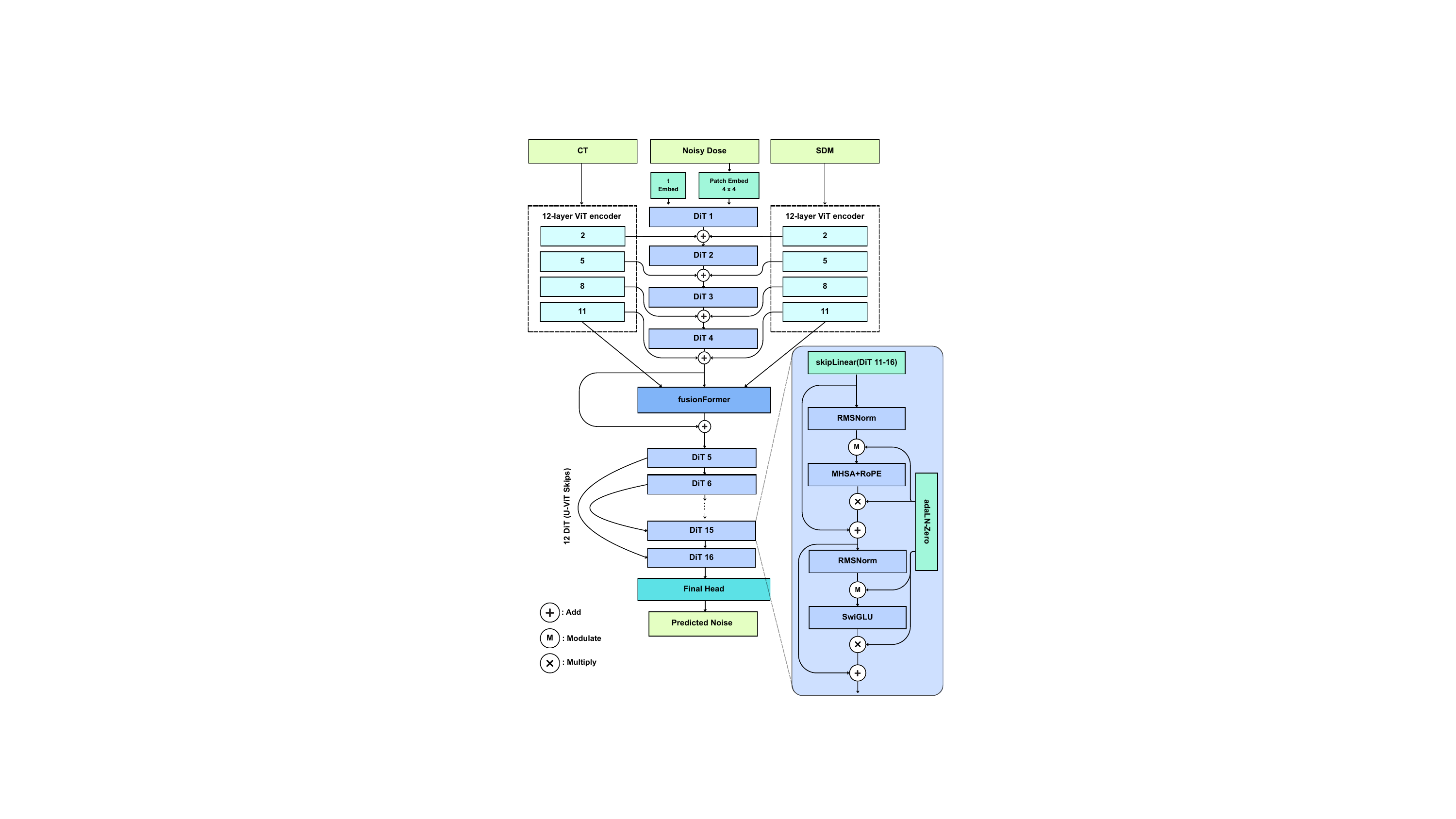}
\caption{\textbf{Architecture of DoseDiT.}
Two 12-layer ViT encoders read the CT and the 20 structure channels (SDM); their layers 2, 5, 8 and 11 are added to the dose tokens after the first four DiT blocks, before a fusion transformer and twelve DiT blocks with U-ViT skip connections.
Right: one DiT block, with RMSNorm, attention with rotary position embeddings, a SwiGLU feed-forward layer and adaLN-Zero modulation by the timestep embedding.}
\label{fig:supp-dit}
\end{figure}
\paragraph{Model source.}
To evaluate DRG on more architectures, we implement DoseDiT on the DoseDiff code base, replacing its UNet denoiser with a diffusion transformer~\cite{peebles2023scalable} (\cref{fig:supp-dit}).
The noisy dose is cut into $4\times4$ patches, a $32\times32$ token grid; two vision-transformer encoders turn the CT, upsampled to $512^2$ and cut into $16\times16$ patches, and the 20 structure channels, cut into $4\times4$ patches, into tokens on the same grid.
Both encoders are built from the ViT-S/16 implementation of DINOv3~\cite{simeoni2025dinov3}.
Their layers 2, 5, 8 and 11 are added to the dose tokens after the first four DiT blocks; a fusion transformer, as in DoseDiff, then takes the last CT features as queries, the last structure features as keys and the dose tokens as values, and adds its output to the dose tokens.
Twelve further DiT blocks with U-ViT long skip connections and a convolutional head predict the noise.

\paragraph{Inputs.}
Those of DoseDiff (\cref{sec:supp-dd}), with the same preprocessing and the same $c_{\setminus k}$.

\paragraph{Training.}
We train from scratch, encoders included, on the 200 training plans with the augmentation of DoseDiff, for 1600 epochs with AdamW (learning rate $10^{-4}$, $10^{-5}$ for the encoders), batch size 24.

\subsection{DoseDiff on GDP-HMM}
\label{sec:supp-gdp}
\paragraph{Model source.}
No public diffusion predictor is available for GDP-HMM~\cite{gao2025gdphmm}, so we train the DoseDiff implementation of \cref{sec:supp-dd} on it, with a 47-channel structure input and $128\times160$ slices.

\paragraph{Inputs.}
The CT and 47 channels, on a $96\times128\times160$ crop around the isocenter: PTV and optimization-shell maps scaled by the prescription, 32 OAR slots, the body mask, beam-geometry maps and four prompt planes (technique, number of PTVs, site, cohort).
The dose is scaled so that the $D_{97}$ of the highest-dose PTV equals its prescription and is mapped from $[0,80]$\,Gy to $[-1,1]$.
For $c_{\setminus k}$, an OAR's slot is set to zero; for a PTV, its PTV and shell maps are set to zero and the PTV-count prompt is reduced by one.

\paragraph{Training.}
We train on the 2727 training plans with AdamW, a step-decayed learning rate starting at $10^{-4}$, batch size 32 and random flips and rotations.

\subsection{DoseDiff on the local cohort}
\label{sec:supp-local}
\paragraph{Model source.}
No public diffusion predictor is available for the local cohort either, so we train the DoseDiff implementation of \cref{sec:supp-dd} on it, with the MR in place of the CT and a 12-channel structure input.

\paragraph{Inputs.}
The MR and 12 channels, on a $128^3$ crop at $1$\,mm spacing around the targets, limited to the axial slices that the MR covers: a target prescription map, masks of the brainstem, cochleae, optic nerves, chiasm and body, and signed distance maps of the targets and these five structures.
MR and dose are mapped to $[-1,1]$, the MR by its 1st and 99.5th percentiles and the dose from $[0,36]$\,Gy.
For $c_{\setminus k}$, the mask and the distance map of structure $k$ (for a target, its prescription map) are set to zero.

\paragraph{Training.}
We train on the 101 training plans for 1200 epochs with AdamW (learning rate $10^{-4}$, divided by 10 after 70\,\% of the epochs), batch size 192 and the augmentation of \cref{sec:supp-gdp}.

\subsection{More implementation details of DRG}
\label{sec:supp-solve}
\begin{table}[t]
\centering
\caption{The two hyperparameters of DRG chosen on validation data, per model: $\gamma$, which sets $\sigma_\beta=0.03\sqrt{\gamma}$, and $\lambda$.}
\label{tab:supp-hparams}
\small
\setlength{\tabcolsep}{4pt}%
\begin{tabular}{@{}lccccc@{}}
\toprule
 & \multicolumn{3}{c}{OpenKBP} & GDP-HMM & Local\\
 & DoseDiff & MD-Dose & DoseDiT & DoseDiff & DoseDiff\\
\midrule
$\gamma$ & 3 & 0.1 & 0.003 & 0.1 & 0.3\\
$\lambda$ & 0.1 & 0.03 & 0.1 & 0 & 1\\
\bottomrule
\end{tabular}
\end{table}

\begin{table}[t]
\centering
\caption{Hyperparameters of the comparison methods: the value chosen on validation cases per setting (DD, MD, DiT, GDP, Local as in \cref{tab:main}).}
\label{tab:supp-grids}
\footnotesize
\setlength{\tabcolsep}{3pt}
\begin{tabular}{@{}llccccc@{}}
\toprule
Method & Param. & DD & MD & DiT & GDP & Local\\
\midrule
DiffPIR & $\zeta$ & 0.1 & 0.3 & 1 & 1 & 1\\
DPS & $\zeta$ & 1 & 1 & 1 & 3 & 0.3\\
TMPD & $\sigma_y$ & 0.1 & 0.1 & 0.3 & 0.3 & 0.3\\
TFG & $\bar\mu$ & 0.25 & 0.0625 & 0.25 & 0 & 0.0625\\
 & $\bar\rho$ & 0.25 & 0.25 & 0.25 & 1 & 0.0625\\
DAPS & lr & $3\cdot10^{-6}$ & $3\cdot10^{-6}$ & $3\cdot10^{-6}$ & $3\cdot10^{-6}$ & $3\cdot10^{-6}$\\
 & $\tau$ & $10^{-3}$ & $10^{-3}$ & $10^{-3}$ & $10^{-2}$ & $10^{-2}$\\
DDiff & $\gamma_0$ & 16 & 16 & 16 & 1.8 & 1.8\\
DNO & lr & $10^{-2}$ & $10^{-2}$ & $10^{-3}$ & $10^{-2}$ & $10^{-3}$\\
FK-8 & $\lambda$ & 100 & 1 & 1 & 10 & 3\\
\bottomrule
\end{tabular}
\end{table}

\begin{table*}[t]
\centering
\caption{\textbf{The errors of \cref{tab:main} in absolute values.}
Cohort-mean errors on the test sets in Gy, lower is better: the target error, the official metric, the DVH score and the outside error (\cref{sec:setup}) of the unsteered prediction and of every method, on the settings of \cref{tab:main}, whose cells are $100\,(1-e/e_{\mathrm{unsteered}})$ of these values before rounding.
Best per column in bold, none in the target-error block because direct edit enforces target error to 0.}
\label{tab:supp-abs}
\footnotesize
\setlength{\tabcolsep}{5pt}
\begin{tabular*}{0.82\textwidth}{@{\extracolsep{\fill}}lcccccccccc}
\toprule
& \multicolumn{5}{c}{Target error} & \multicolumn{5}{c}{Official metric} \\
\cmidrule(lr){2-6} \cmidrule(lr){7-11}
Method & DD & MD & DiT & GDP & Local & DD & MD & DiT & GDP & Local \\
\midrule
Unsteered & 1.417 & 2.494 & 1.415 & 1.039 & 0.555 & 2.707 & 4.308 & 2.756 & 2.508 & 0.710 \\
\midrule
\multicolumn{11}{c}{\textit{Clean-space correction and alternating inference}} \\
\midrule
Direct edit & 0.000 & 0.000 & 0.000 & 0.000 & 0.000 & 2.685 & 4.332 & 2.728 & 2.419 & 0.688 \\
DDNM~\cite{wang2023ddnm} & 0.001 & 0.002 & 0.001 & 0.000 & 0.000 & 2.680 & 4.322 & 2.684 & 2.284 & 0.694 \\
DiffPIR~\cite{zhu2023diffpir} & 0.001 & 0.003 & 0.001 & 0.000 & 0.000 & 2.677 & 4.321 & 2.683 & 2.263 & 0.693 \\
DAPS~\cite{zhang2025daps} & 1.281 & 2.237 & 1.219 & 1.026 & 0.560 & 2.780 & 4.434 & 2.809 & 2.560 & 0.769 \\
DDiff~\cite{kim2026ddiff} & 0.993 & 1.727 & 0.865 & 0.875 & 0.484 & 2.707 & 4.251 & 2.700 & 2.433 & 0.705 \\
\midrule
\multicolumn{11}{c}{\textit{Gradient-based guidance}} \\
\midrule
DPS~\cite{chung2023dps} & 1.028 & 1.605 & 0.984 & 1.006 & 0.487 & 2.701 & 4.264 & 2.721 & 2.387 & 0.711 \\
$\Pi$GDM~\cite{song2023pigdm} & 0.623 & 0.916 & 0.432 & 0.219 & 0.311 & 2.688 & 4.252 & 2.702 & 2.278 & 0.694 \\
TMPD~\cite{boys2024tmpd} & 1.119 & 0.824 & 0.954 & 0.501 & 0.535 & 2.685 & 4.150 & 2.700 & 2.334 & 0.709 \\
TFG~\cite{ye2024tfg} & 0.797 & 1.631 & 0.829 & 0.849 & 0.321 & 2.693 & 4.256 & 2.710 & 2.396 & 0.710 \\
\midrule
\multicolumn{11}{c}{\textit{Noise optimization and sample selection}} \\
\midrule
DNO~\cite{tang2025dno} & 1.153 & 1.216 & 0.994 & 0.165 & 0.455 & 2.698 & 4.158 & 2.692 & 2.252 & 0.704 \\
Best-of-8 & 1.397 & 2.430 & 1.353 & 0.967 & 0.536 & 2.703 & 4.273 & 2.745 & 2.492 & 0.708 \\
FK-8~\cite{singhal2025fk} & 1.404 & 2.439 & 1.370 & 0.939 & 0.537 & 2.711 & 4.294 & 2.733 & 2.409 & 0.711 \\
\midrule
DRG (ours) & 0.016 & 0.031 & 0.019 & 0.002 & 0.014 & $\mathbf{2.620}$ & $\mathbf{4.094}$ & $\mathbf{2.619}$ & $\mathbf{2.231}$ & $\mathbf{0.606}$ \\
\bottomrule
\end{tabular*}

\vspace{6pt}

\setlength{\tabcolsep}{5pt}
\begin{tabular*}{0.82\textwidth}{@{\extracolsep{\fill}}lcccccccccc}
\toprule
& \multicolumn{5}{c}{DVH score} & \multicolumn{5}{c}{Outside error} \\
\cmidrule(lr){2-6} \cmidrule(lr){7-11}
Method & DD & MD & DiT & GDP & Local & DD & MD & DiT & GDP & Local \\
\midrule
Unsteered & 1.566 & 2.660 & 1.597 & 1.215 & 0.899 & 2.928 & 4.679 & 2.981 & 2.831 & 0.623 \\
\midrule
\multicolumn{11}{c}{\textit{Clean-space correction and alternating inference}} \\
\midrule
Direct edit & 0.931 & 1.519 & 0.902 & 0.683 & $\mathbf{0.497}$ & 2.928 & 4.679 & 2.981 & 2.831 & 0.623 \\
DDNM~\cite{wang2023ddnm} & 0.902 & 1.539 & 0.864 & 0.607 & 0.518 & 2.917 & 4.646 & 2.929 & 2.713 & 0.636 \\
DiffPIR~\cite{zhu2023diffpir} & 0.901 & 1.547 & 0.873 & 0.618 & 0.511 & 2.914 & 4.646 & 2.930 & 2.690 & 0.634 \\
DAPS~\cite{zhang2025daps} & 1.476 & 2.561 & 1.474 & 1.179 & 0.853 & 3.006 & 4.799 & 3.032 & 2.891 & 0.655 \\
DDiff~\cite{kim2026ddiff} & 1.444 & 2.387 & 1.427 & 1.092 & 0.834 & 2.939 & 4.637 & 2.932 & 2.790 & 0.615 \\
\midrule
\multicolumn{11}{c}{\textit{Gradient-based guidance}} \\
\midrule
DPS~\cite{chung2023dps} & 1.401 & 2.359 & 1.420 & 1.322 & 0.840 & 2.931 & 4.658 & 2.954 & 2.759 & 0.629 \\
$\Pi$GDM~\cite{song2023pigdm} & 1.133 & 1.723 & 1.056 & 0.725 & 0.710 & 2.928 & 4.663 & 2.947 & 2.704 & 0.624 \\
TMPD~\cite{boys2024tmpd} & 1.451 & 1.694 & 1.321 & 0.863 & 0.889 & 2.910 & 4.545 & 2.935 & 2.715 & 0.627 \\
TFG~\cite{ye2024tfg} & 1.485 & 2.472 & 1.557 & 1.161 & 0.744 & 2.929 & 4.660 & 2.951 & 2.764 & 0.629 \\
\midrule
\multicolumn{11}{c}{\textit{Noise optimization and sample selection}} \\
\midrule
DNO~\cite{tang2025dno} & 1.404 & 1.829 & 1.309 & 0.674 & 0.825 & 2.926 & 4.532 & 2.929 & 2.656 & 0.624 \\
Best-of-8 & 1.554 & 2.622 & 1.552 & 1.175 & 0.890 & 2.924 & 4.637 & 2.970 & 2.816 & 0.622 \\
FK-8~\cite{singhal2025fk} & 1.556 & 2.633 & 1.560 & 1.140 & 0.887 & 2.934 & 4.663 & 2.956 & 2.772 & 0.629 \\
\midrule
DRG (ours) & $\mathbf{0.785}$ & $\mathbf{1.399}$ & $\mathbf{0.763}$ & $\mathbf{0.560}$ & 0.518 & $\mathbf{2.861}$ & $\mathbf{4.459}$ & $\mathbf{2.869}$ & $\mathbf{2.647}$ & $\mathbf{0.541}$ \\
\bottomrule
\end{tabular*}
\end{table*}

\begin{figure}[t]
\centering
\includegraphics[width=\linewidth]{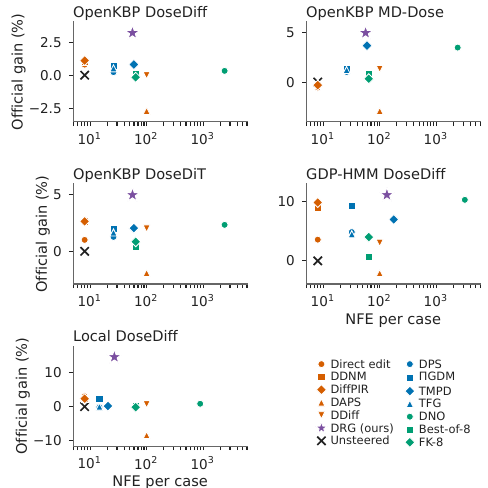}
\caption{\textbf{Official gain against compute.}
Gain in the official metric (\cref{tab:main}) against the number of network function evaluations (NFE) per case, on a log scale; color: method category, marker: method, as in \cref{fig:tension}; cross: unsteered prediction.
One evaluation is a network pass over all slices of a case; a pass over part of the slices counts pro rata and a backward pass as two.}
\label{fig:compute}
\end{figure}
\paragraph{Reduced context.}
The reduced context $c_{\setminus k}$ sets the channels of structure $k$ to zero (\cref{sec:prelim}), as specified per predictor in \cref{sec:supp-dd,sec:supp-md,sec:supp-dit,sec:supp-gdp,sec:supp-local}.
All five predictors encode a structure that a plan lacks by all-zero channels, as the preprocessing of DoseDiff~\cite{zhang2024dosediff} does, so $c_{\setminus k}$ presents structure $k$ as absent.
\Cref{prop:tilt,prop:responses,prop:surface} also hold with this input as the reference context (\cref{sec:supp-tilt}), so $c_{\setminus k}$ need not marginalize structure $k$ out.
Plans without a given structure are common in training, a structure dropout that the data provide: on OpenKBP, every OAR and PTV except PTV70 is missing from 7 to 119 of the 200 training plans, and on the local cohort every OAR from 41 to 94 of the 101.
For structures present in every training plan, such as PTV70, $c_{\setminus k}$ is an extrapolation, which is why \cref{sec:prelim} treats it as an approximation.
The context response enters the correction only through the solve, which weighs it by its measured effect on the targets and by its price: on GDP-HMM, where the highest-dose PTV is filled in every training plan and many OAR slots in nearly every one, DRG relies mostly on the noise lever (\cref{fig:shares}).

\paragraph{Solve.}
Up to a factor $\frac12$ and a constant, the price $\mathcal{P}$ is the negative log-density of a Gaussian prior on the strengths, with variances $v_k/\lVert G^N_{:,k}\rVert^2$ for $\alpha_k$ and $\sigma_\beta^2$ for $\beta_k$.
A push of one prior standard deviation changes the metrics by a vector of squared norm $v_k$ for $\alpha_k$ and $\sigma_\beta^2\lVert G^C_{:,k}\rVert^2$ for $\beta_k$; the damping $\mu$ of \cref{eq:mu} is $10^{-3}$ times the mean of these squared norms, so it scales with the responses.
We solve \cref{eq:solve} by bounded-variable least squares, so the box is part of the solve rather than a clip applied afterwards, and apply the full $\Delta$ without a step size.
Where the finite difference gives $v_k\le0$, we set $v_k=0$; all prior variances of the $\alpha_k$ are then increased by $10^{-6}$ times their mean (at least by $10^{-36}$). This floor keeps them positive and makes the noise push of such a structure negligible rather than exactly zero.

\subsection{Hyperparameters of DRG}
\label{sec:supp-hparams}
All models share $B=3$, $h=0.02$ (model units), the 10\,\% PTV cold set, the factor $10^{-3}$ in $\mu$ and the steering steps $\mathcal{S}=\{3,2,1,0\}$.
Two hyperparameters are chosen per model on validation cases (\cref{tab:supp-hparams}): the weight $\lambda$ of the shape term ($\lambda=0$: no shape term) and the scale $\sigma_\beta$ of the context lever.
$\sigma_\beta$, the prior standard deviation of $\beta_k$, sets the exchange rate between the levers: a weight change of $\sigma_\beta$ costs as much as a noise push that moves the metrics by one posterior standard deviation.
Unlike $v_k$, which the noise response measures, the model gives no estimate of this scale, so we search $\sigma_\beta=0.03\sqrt{\gamma}$ over $\gamma\in\{0.003,0.01,0.03,\dots,30\}$, around a base scale from a coarse validation scan.
We choose both on the validation cases by the official metric, among the values whose DVH score is not worse than that of the unsteered prediction.

\subsection{Comparison methods}
\label{sec:supp-baselines}
\begin{figure*}[t]
\centering
\includegraphics[width=\textwidth]{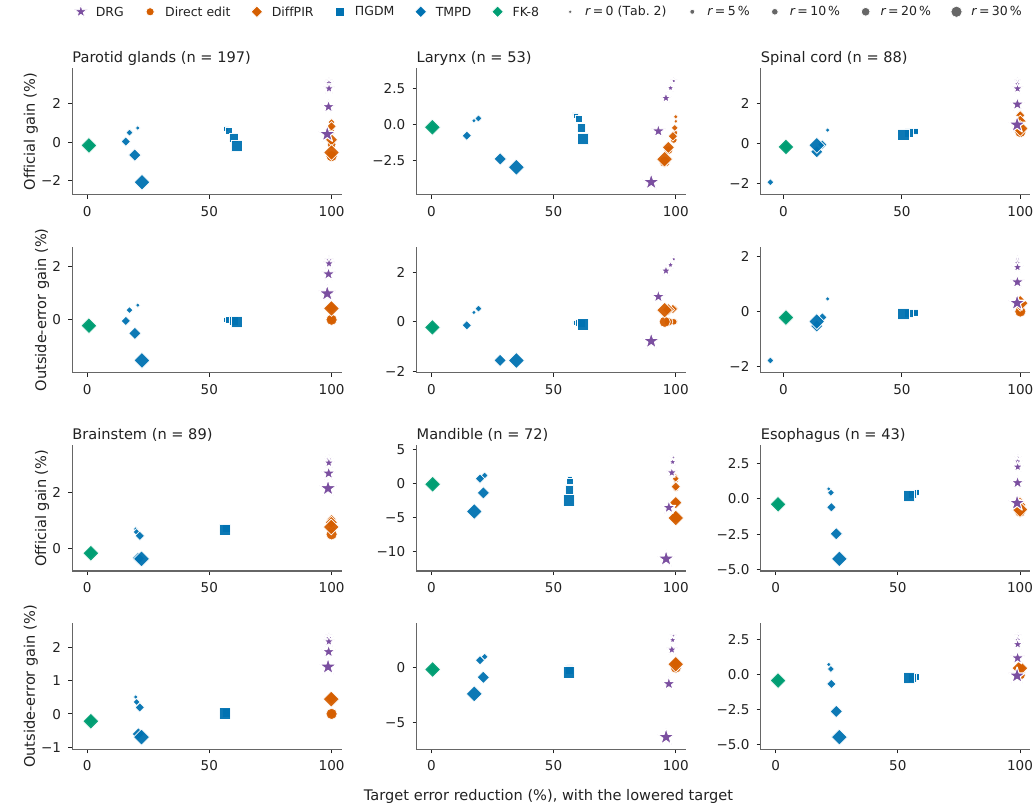}
\caption{\textbf{Robustness to counterfactual targets on OpenKBP with DoseDiff.}
One OAR's mean-dose target is lowered by $r$ (marker size; $r=0$: the clinical targets of \cref{tab:main}, on the cases with that OAR), every other target is clinical; panel titles: the lowered OAR and its number of organs.
Horizontal: reduction of the target error, measured against the lowered target (higher is better).
Vertical: gain in the official metric (first and third rows) and in the outside error (second and fourth rows), both measured against the original clinical dose: they show how far a method moves from the original plan.
All values are relative to the unsteered prediction, in \%.}
\label{fig:supp-cf}
\end{figure*}
All methods start from the initial noise $x_{T-1}$ of the unsteered prediction, and the published methods follow their algorithms as published.
Each method keeps the configuration recommended by its authors wherever our setting allows, so that no method is re-tuned at length for our predictors; only the hyperparameters in \cref{tab:supp-grids}, mostly a guidance strength or step size, are chosen, on the same validation cases and with the same rule as for DRG (\cref{sec:supp-hparams}).

\section{Additional results and discussions}
\label{sec:supp-results}

\subsection{Absolute errors}
\label{sec:supp-abs}
\Cref{tab:supp-abs} lists the errors of \cref{tab:main} in Gy, with the unsteered prediction in the first row.
As in percent, DRG has the lowest official and outside errors on every setting and the lowest DVH score on four.

\subsection{Compute}
\label{sec:supp-compute}
\Cref{fig:compute} plots the official gain against the number of network function evaluations (NFE) per case.
Because each corrected step probes the predictor for its own responses, once per steered structure without that structure's channels and twice on the structure's slices for the noise response (\cref{eq:noise}), DRG needs more evaluations than direct projection methods such as DDNM and DiffPIR, though fewer than DNO.
This is the cost of the method; in return, DRG has the highest official gain on every setting, and since the count grows with the number of steered structures, it falls when physicians set targets for only some of them.

\subsection{The noise lever and TMPD}
\label{sec:supp-noise}
The noise lever is mathematically related to guidance methods that condition the clean estimate on measurements through the posterior covariance, such as TMPD: under (A0), its response $J\mathbf{1}_k$ is, up to $\kappa_t$, the covariance of structure $k$'s dose with the dose everywhere (\cref{sec:levers}).
The two nevertheless differ in what they assume about $J$ and in how they turn it into a correction.
TMPD conditions the clean estimate on the targets as if the posterior were Gaussian with covariance $\kappa_tJ$ and the targets were observed with one noise level $\sigma_y$.
This update needs $\kappa_tJ$ to be a valid covariance, symmetric and positive semi-definite, which holds under (A0) (\cref{lem:tweedie}) but not reliably on the trained predictors: the variance $v_k$ measured from the noise response is non-positive for $2$ to $24\,\%$ of the steered structures over the steering steps, up to $57\,\%$ at $t=3$ on MD-Dose, and almost never at the last step.
The noise lever does not rely on this property: it uses the measured responses in a priced least-squares solve that takes only the scale $v_k$ of the price from (A0) and, with the floor of \cref{sec:supp-solve}, has a unique minimizer that does not increase the linearized residual (\cref{prop:solve}).
In this solve, the price measures the push on each structure in units of that structure's own posterior variance $v_k$ (\cref{eq:price,eq:vk}) rather than with one noise level for all targets, a structure with $v_k\le0$ gets almost no noise push, the shape term favors corrections that shift the dose within each structure rather than reshape it, and the damping is set from the responses themselves (\cref{eq:mu}).

These components belong to DRG, so the noise lever is not a separate method.
The noise-lever row of \cref{tab:ablation} keeps the noise responses $N_k$ (\cref{eq:noise}), the solve with all of its components, the steering steps, the sampler, the targets and the initial noise, and only sets the context strengths $\beta_k$ to zero.
Its strength therefore credits components that DRG introduces: with them, the noise response alone removes $96.2$ to $99.8\,\%$ of the target error and gains $1.1$ to $13.1\,\%$ in the official metric.
The noise lever is also only one of DRG's two levers, and the context lever complements it where the noise response is weak (\cref{fig:shares}), for example on the structures that the floor leaves without a noise push; on OpenKBP DoseDiff, adding the context lever raises the official gain from $1.1$ to $3.2\,\%$.
With both levers, DRG removes at least as much of the target error as either lever alone on every setting; where the noise lever alone has the higher official gain, on MD-Dose and DoseDiT ($p=0.015$ and $0.97$), DRG removes more of the target error ($98.8$ against $97.7\,\%$ and $98.7$ against $98.5\,\%$, $p<10^{-3}$).
DRG measures the noise and the context response of every steered structure and solves for both strengths jointly, so it needs no choice of lever per predictor, and it averages a higher official gain than the better single lever of each setting (\cref{tab:ablation}).

\subsection{Robustness to counterfactual targets}
\label{sec:supp-cf}
The protocol of \cref{sec:setup} derives the targets from the clinical dose $x^\star$, which makes $x^\star$ a ground truth that meets every target; once a target is moved away from $q_k(x^\star)$, meeting it requires redistributing dose, and no such ground truth exists.
We therefore use moved targets only to test robustness: we lower the mean-dose target of one OAR at a time to $(1-r)$ times its clinical value, $r\in\{5,10,20,30\,\%\}$, keep the other targets clinical, and rerun DRG and five baselines with DoseDiff on the OpenKBP test cases (\cref{fig:supp-cf}).
DRG still removes $90$ to $99\,\%$ of the error to the lowered targets for every OAR and every $r$; among the five baselines, only the direct edit and DiffPIR remove a similar share.
Since $x^\star$ remains the reference, the official and outside errors now measure how far each method moves away from the original plan.
When a target is lowered by at most $10\,\%$, DRG has the highest official and outside gains on all seven OARs: it follows the lowered target while changing the original plan the least, which indicates robustness to moved targets.
Another method comes closer to the original plan only on some OARs at the two largest reductions ($20$ and $30\,\%$).

\end{document}